\documentclass[10pt,twocolumn,letterpaper]{article}

\usepackage{cvpr}
\usepackage{xcolor}
\usepackage{authblk}
\usepackage{times}
\usepackage{epsfig}
\usepackage{graphicx}
\usepackage{amsmath}
\usepackage{amssymb}
\usepackage{subcaption}
\usepackage{booktabs}
\usepackage{colortbl}
\usepackage{appendix}

\usepackage{comment}

\usepackage[noadjust]{cite}

\usepackage{tikz}
\usetikzlibrary{calc}
\usepackage{multirow}
\usepackage{enumitem}

\usepackage{pifont}% http://ctan.org/pkg/pifont

\definecolor{cadmiumgreen}{rgb}{0.0, 0.42, 0.24}
\newcommand{\cmark}{\scalebox{1.5}{\textcolor{cadmiumgreen}{\ding{51}}}}%
\newcommand{\xmark}{\scalebox{1.5}{\textcolor{red}{\ding{55}}}}%

\usepackage[pagebackref=true,breaklinks=true,colorlinks,bookmarks=false]{hyperref}

\cvprfinalcopy % *** Uncomment this line for the final submission

\def\cvprPaperID{4934} % *** Enter the CVPR Paper ID here

\ifcvprfinal\fi

\newcommand{\arch}{EVRNet}

\begin{document}

%%%%%%%%% TITLE
\title{\arch: Efficient Video Restoration on Edge Devices}

\author[1]{Sachin Mehta\thanks{This work was done as a part of internship at Facebook.}}
\author[2]{Amit Kumar}
\author[2]{Fitsum Reda}
\author[2]{Varun Nasery}
\author[2]{Vikram Mulukutla}
\author[2]{Rakesh Ranjan}
\author[2]{Vikas Chandra}
\affil[1]{University of Washington \quad $^2$ Facebook Inc.}
%\affil[2]{}
%\affil[ ]{\it {\{sacmehta\}@cs.washington.edu}\quad   mohammadr@allenai.org}

\maketitle

%%%%%%%%% ABSTRACT
\begin{abstract}
    Video transmission applications (e.g., conferencing) are gaining momentum, especially in times of global health pandemic. Video signals are transmitted over lossy channels, resulting in low-quality received signals. To restore videos on recipient edge devices in real-time, we introduce an efficient video restoration network, \arch. 
    \arch~efficiently allocates parameters inside the network using alignment, differential, and fusion modules. With extensive experiments on video restoration tasks (deblocking, denoising, and super-resolution), we demonstrate that \arch~delivers competitive performance to existing methods with significantly fewer parameters and MACs. For example, \arch~has $260\times$ fewer parameters and $958\times$ fewer MACs than enhanced deformable convolution-based video restoration network (EDVR) for $4\times$ video super-resolution while its SSIM score is 0.018 less than EDVR. We also evaluated the performance of \arch~under multiple distortions on unseen dataset to demonstrate its ability in modeling variable-length sequences under both camera and object motion.
\end{abstract}

%%%%%%%%% BODY TEXT
\section{Introduction}
\begin{figure}[b!]
    \centering
    \begin{subfigure}[b]{\columnwidth}
        \centering
        \includegraphics[width=0.9\columnwidth]{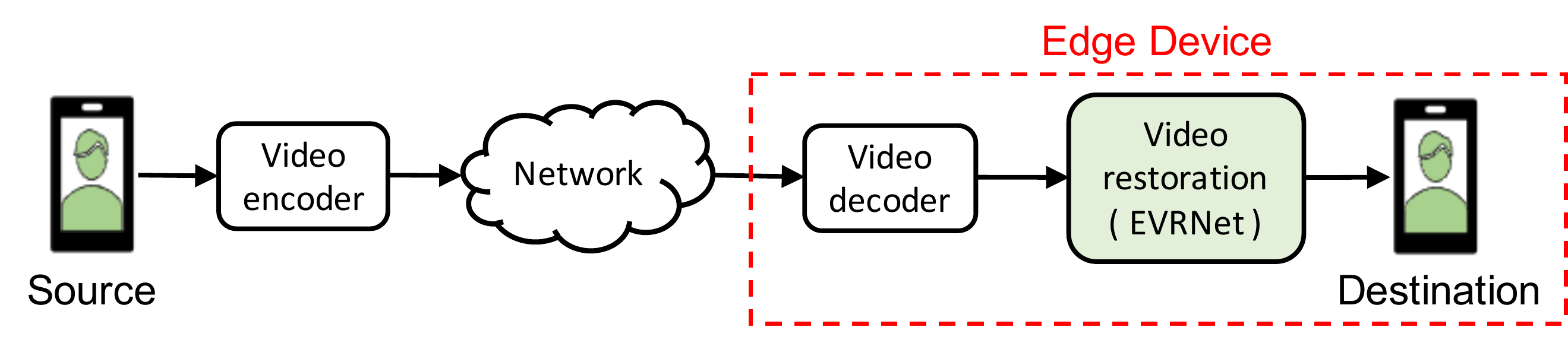}
        \caption{\arch~in video conferencing application.}
        \label{fig:video_transmission_system}
    \end{subfigure}
    \vfill
    \begin{subfigure}[b]{\columnwidth}
        \centering
        \begin{tabular}{c}
             \includegraphics[width=0.9\columnwidth]{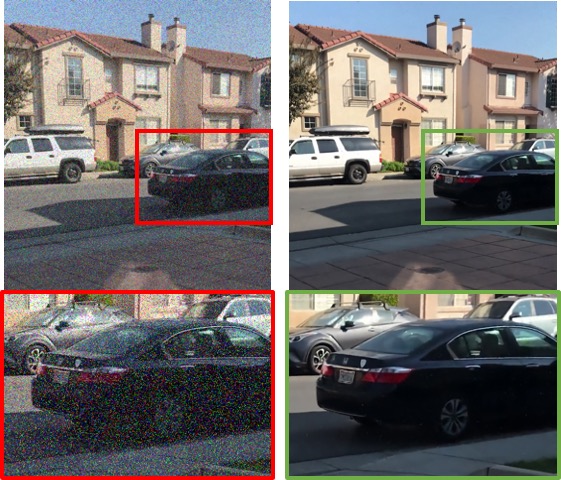} \\
             \vspace{0.5mm}
             \\
            \includegraphics[width=0.9\columnwidth]{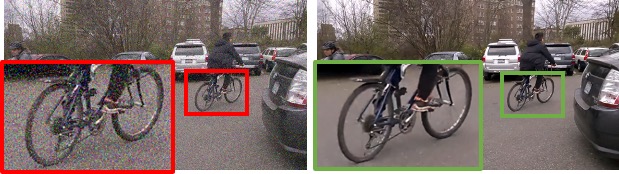}
        \end{tabular}
        \caption{Sample \arch~results on unseen videos. \textbf{Left:} compressed and noisy frames. \textbf{Right:} Restored frames.}
        \label{fig:taser_examples}
    \end{subfigure}
    \caption{\textbf{\arch~on edge devices.} \textbf{(a)} shows how \arch~is integrated to an edge device while \textbf{(b)} shows the results of \arch~on H264 compressed and noisy (Gaussian + salt and pepper) ``unseen" videos. \arch~is able to restore the videos with multiple artifacts. See Appendix \ref{ssec:qual_ablate} for more results.}
\end{figure}
\begin{figure*}[t!]
    \centering
    \includegraphics[width=1.8\columnwidth]{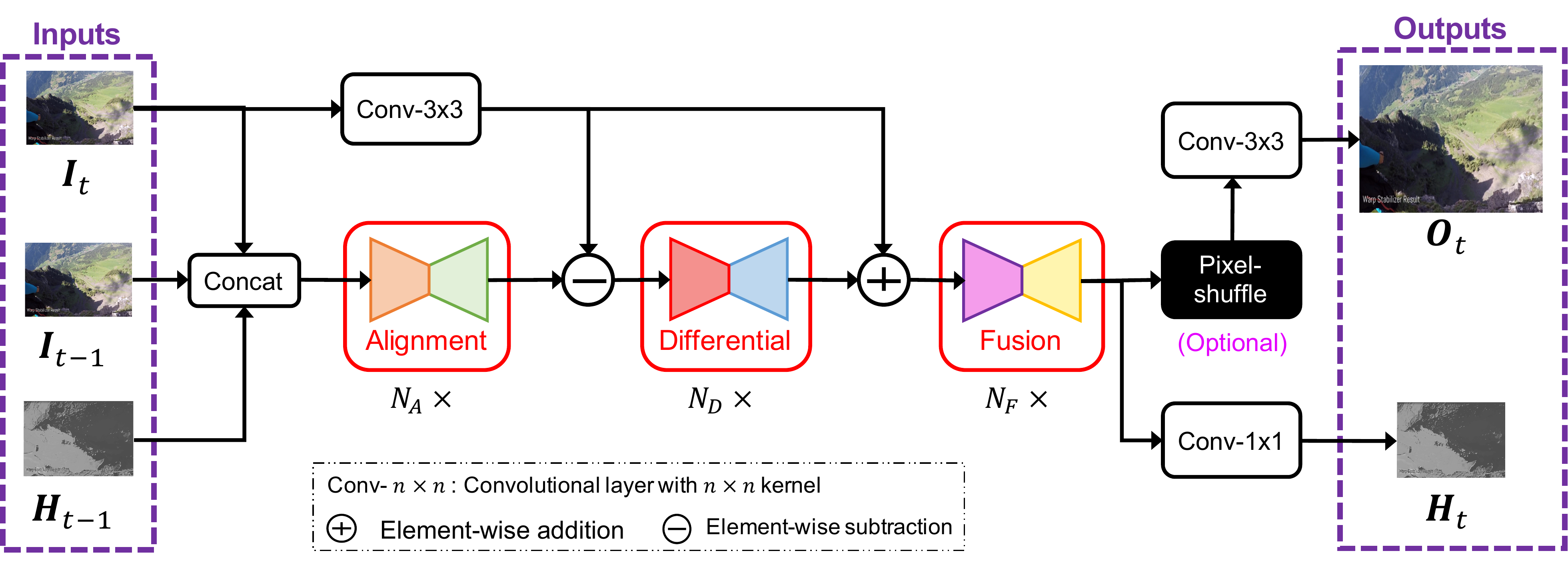}
    \caption{\textbf{Overview of the \arch~architecture for videos}. The \arch~architecture takes the current RGB $\mathbf{I}_t$, previous RGB $\mathbf{I}_{t-1}$, and previous latent $\mathbf{H}_{t-1}$ frames as inputs and produces two outputs: restored RGB frame $\mathbf{O}_t$ and latent frame $\mathbf{H}_t$. The pixel-shuffle operation is optional and is used only for super-resolution tasks. The alignment, differential, and fusion modules are light-weight and efficient encoder-decoder networks (see Figure \ref{fig:align_diff_fuse}) with $N_A$, $N_D$, and $N_F$ layers, respectively.}
    \label{fig:esrnet_arch}
\end{figure*}

Video restoration aims at recovering the expected quality of videos in recipient devices. Deep neural network-based solutions \cite{edvr2019wang,xue2019video,zhang2017beyond,krull2019noise2void,haris2019recurrent} achieve high accuracy on these tasks, but they are computationally very expensive. For example, a deformable convolution-based video restoration network, EDVR \cite{edvr2019wang}, has 21.1 million parameters and requires 9.96 TMACs (multiplication-addition operations) for up-sampling a 360p video frame by a factor of 4. Many video transmission applications (e.g., video streaming and video conferencing) run on edge devices, such as smartphones. The trend is likely to continue with the on-going global health pandemic and the need for remote and virtual collaboration. Edge devices have limited computational resources, memory, and energy. As such, heavy networks are not suitable for edge devices. Additionally, video signals at source often undergo lossy compression for transmission under limited network bandwidth (see Figure \ref{fig:video_transmission_system}). Because of compression and transmission noise, the quality of received video signals is low. In order to be effective, these applications should be able to restore high quality and temporally stable videos with low latency on edge devices.

In this work, we propose an efficient and unified neural network (see Figure \ref{fig:esrnet_arch}) that restores videos with high quality on edge devices in real-time. Efficient Video Restoration Network, (\arch), is inspired by traditional computer vision methods for motion estimation and image enhancement \cite{lucas1981iterative,polesel2000image, deng2010generalized}. Briefly, \arch~uses an alignment module to align current and previous frames without optical flow. High-frequency components (e.g., object edges) are often lost during compression. To restore such details, \arch~uses a differential and fusion module. The differential module learns representations corresponding to high-frequency components while the fusion module uses these representations along with the input to produce high-quality output (see Figure \ref{fig:taser_examples}). \arch~more efficiently allocates parameters and operations inside each of these modules using small and light-weight encoder-decoder networks.

We evaluate \arch's performance on large scale Vimeo-90K dataset \cite{xue2019video} on three restoration tasks: (1) deblocking, (2) denoising, and (3) super-resolution. \arch~delivers competitive performance as state-of-the-art methods but with significantly fewer parameters and MACs. For example, on the task of video deblocking and denoising, \arch~delivers similar performance to ToFlow \cite{xue2019video} but with $46\times$ and $13.63 \times$ fewer MACs and parameters, respectively. On the task of $4\times$ video super-resolution, \arch~has slightly lower SSIM score (0.018) than EDVR \cite{edvr2019wang}, but has $260\times$ fewer parameters and $958\times$ fewer MACs. 

To summarize, the main contributions of this paper are:
\begin{itemize}
  \item A novel efficient video restoration network capable of running at real-time on edge devices.
  \item A unified neural network, \arch, that jointly removes compression and noise artifacts that are prevalent in video transmission pipeline. 
  \item Qualitative and quantitative results along with comparisons with state-of-the-art methods on three video restoration tasks, demonstrating \arch's competitive performance, while having significantly fewer network parameters and MACs.
\end{itemize}

\begin{figure*}[t!]
    \centering
    \begin{subfigure}[b]{0.58\columnwidth}
         \centering
         \includegraphics[height=140px]{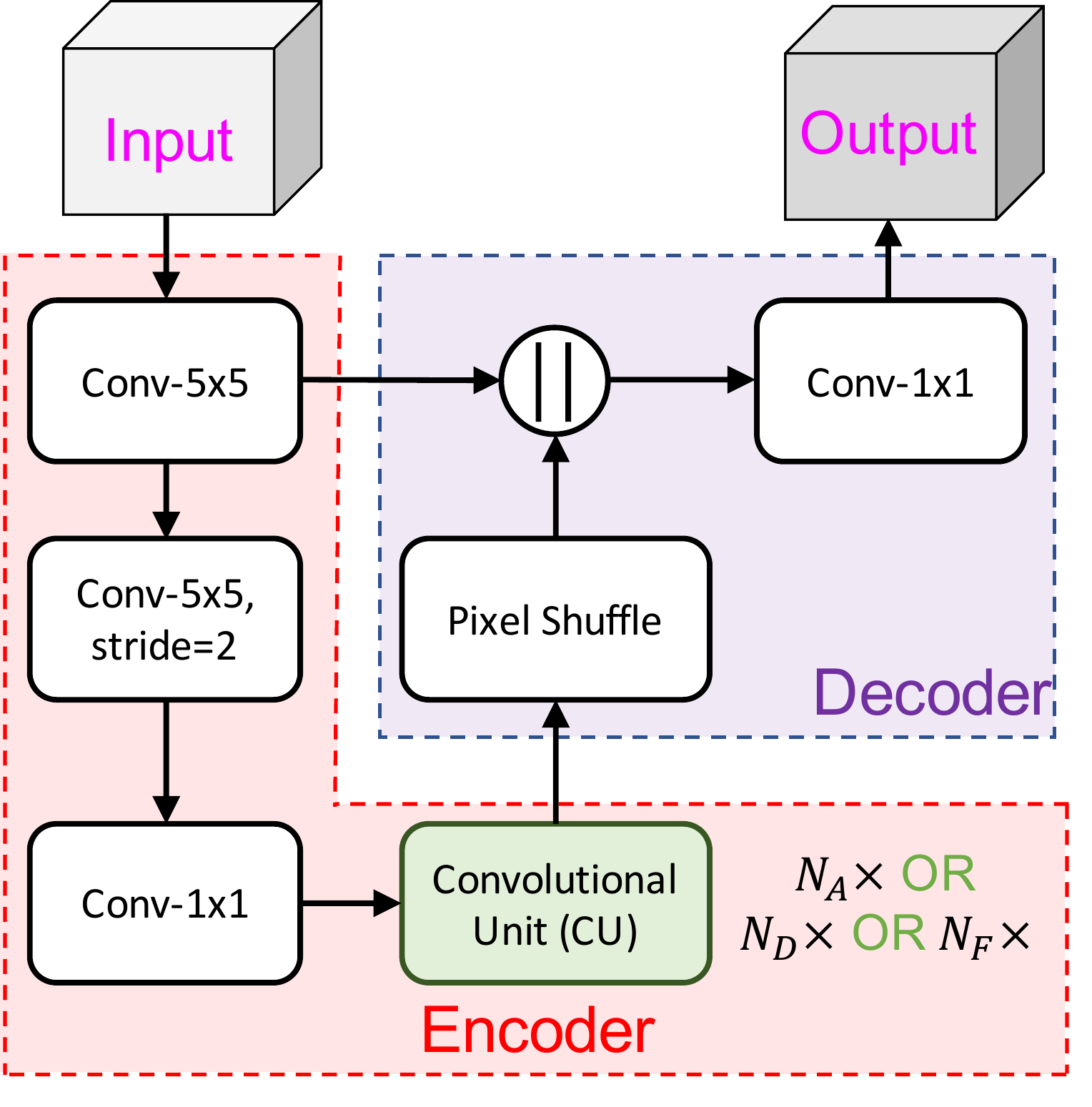}
         \caption{Small encoder-decoder network}
         \label{fig:tiny_enc_dec}
     \end{subfigure}
     \hfill
     \begin{subfigure}[b]{1.4\columnwidth}
         \centering
         \includegraphics[height=140px]{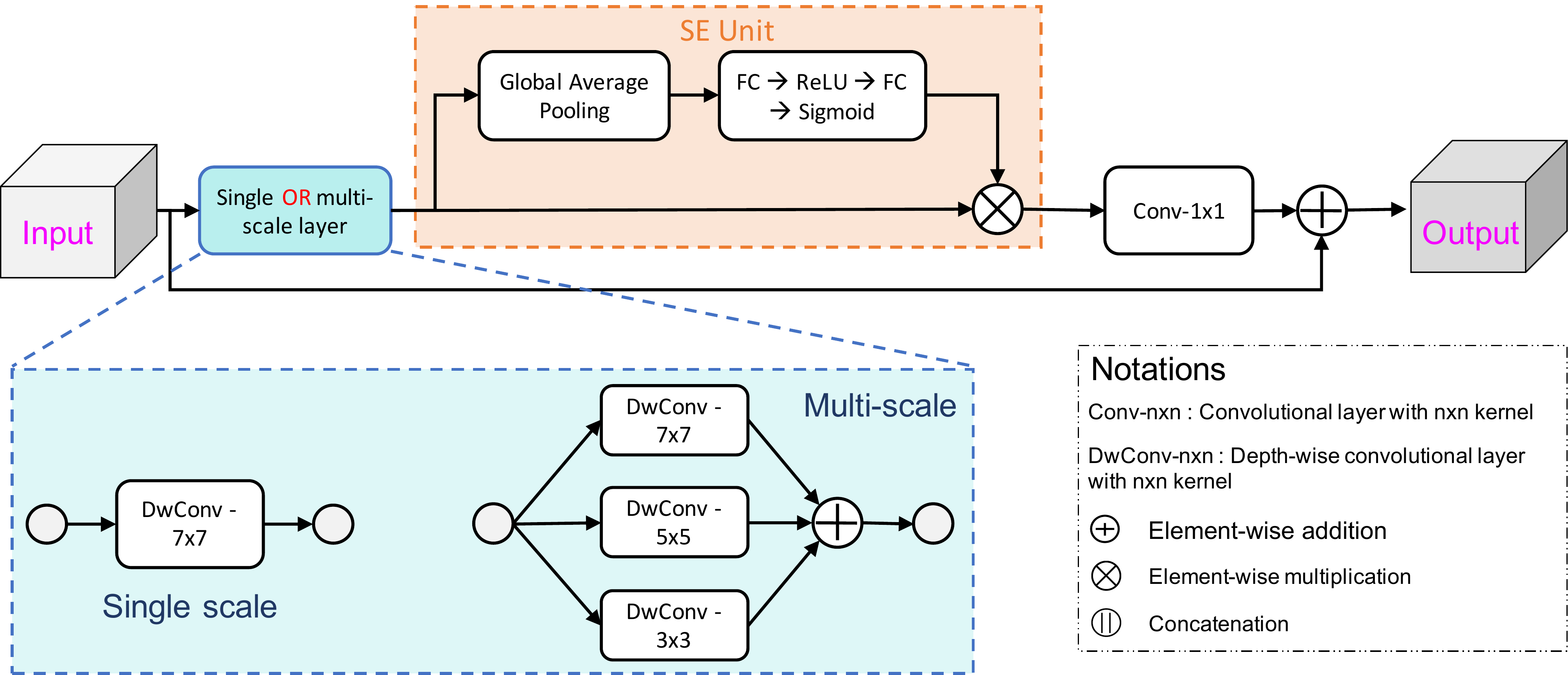}
         \caption{Convolutional unit (CU)}
         \label{fig:conv_units}
     \end{subfigure}
    \caption{\textbf{Overview of alignment, differential, and fusion module}. Each of these modules are identical in construction, i.e., they follow an encoder-decoder structure \textbf{(a)}, with an exception to the number of convolutional units \textbf{(b)}. The alignment, differential, and fusion module stacks $N_A$, $N_D$, and $N_F$ convolutional units (CUs) to learn deep representations, respectively.}
    \label{fig:align_diff_fuse}
\end{figure*}

\section{Related Work}
\label{sec:literature_revie}
Designing deep neural networks for image and video restoration tasks is an active area of research. In this section, we first briefly review these approaches followed by efforts in improving the efficiency of neural networks.

\vspace{1mm}
\noindent \textbf{Image and video restoration:} Video deblocking (e.g., \cite{dong2015compression, zhang2017beyond, maggioni2012video, xue2019video, lu2018deep}), video denoising (e.g., \cite{xue2019video, maggioni2012video, zhang2017beyond, krull2019noise2void, yu2020joint}), and super-resolution (e.g., \cite{dong2014learning, ledig2017photo, wang2018esrgan, kim2016accurate, tong2017image, huang2015bidirectional, caballero2017real,liu2017robust,sajjadi2018frame,tao2017detail,edvr2019wang}) are three main video restoration tasks that have been studied widely in the literature. Video deblocking aims at removing artifacts that arises due to image or video compression (e.g., checkerboard patterns). Video denoising aims at removing noise-related artifacts that may arise due to noisy transmission channel (e.g., Internet). Super-resolution aims at producing a high-resolution images/videos from low-resolution images/videos. Most methods are studied on one of these tasks and are computationally very expensive. For example, ToFlow \cite{xue2019video} has about 466 GMACs for denoising (or deblocking) a 360p video. Also, some video restoration methods use optical flow  (e.g., \cite{xue2019video, caballero2017real, bao2019memc}) which is computed using deep flow networks, such as FlowNet \cite{dosovitskiy2015flownet, IMKDB17}, PWCNet \cite{Sun_CVPR_2018}, and SpyNet \cite{ranjan2017optical}). Computing optical flow with these networks is expensive and this limits the practical applicability of such approaches, especially on resource-constrained devices (e.g., Smartphones). Similar to \cite{edvr2019wang, jo2018deep, tian2020tdan}, \arch~also does implicit alignment between consecutive frames using the pyramidal structure in the alignment module and handles large motion without optical flow. Importantly, \arch~can restore videos with high-quality in real-time on edge devices.

\vspace{1mm}
\noindent \textbf{Efficient networks:} Efficient deep neural networks, an active area in both academic and industrial research, aims at reducing the network parameters and MACs by designing efficient learnable layers (e.g., depth-wise convolutions \cite{chollet2017xception}) or quantization or compression or pruning. The most similar to our work are the methods on architecture design (hand-crafted \cite{howard2017mobilenets, sandler2018mobilenetv2, ma2018shufflenet, mehta2019espnetv2} and learned \cite{zoph2016neural,tan2019mnasnet,howard2019searching,tan2019efficientnet}). Similar to these methods, \arch~also uses depth-wise convolutions for learning representations efficiently. Network compression \& pruning (e.g., \cite{han2015deep,wen2016learning,li2018constrained,he2018amc, yu2018nisp,  molchanov2019importance}), quantization (e.g., \cite{rastegari2016xnor,wu2016quantized,courbariaux2016binarized,andri2018yodann}), and distillation (e.g., \cite{hinton2015distilling,gupta2016cross,yim2017gift}) are important complementary efforts that can be further used to improve the efficiency of \arch.

\section{\arch}
\label{sec:esrnet}
We propose \arch, an \textbf{E}fficient \textbf{V}ideo \textbf{R}estoration \textbf{Net}work, to remove artifacts and restore videos in edge devices in real-time (schematic shown in Figure \ref{fig:esrnet_arch}). \arch~takes inspirations from traditional techniques in motion estimation and image enhancement \cite{lucas1981iterative, polesel2000image}. Specifically, \arch~uses an alignment module  based on a pyramidal structure to model the motion without explicit use of optical flow. To restore high-frequency details (e.g., edges) that may be lost due to distortions (e.g., compression), \arch~uses differential and fusion module. These modules learn high-frequency components which are then added back to achieve sharp details. Following sub-sections describe the overall architecture of \arch~in detail.

\subsection{\arch~Architecture}
\label{ssec:esrnet_arch}
\arch~is an auto-regressive network that efficiently models the relationships between a current frame $\mathbf{I}_t \in \mathbb{R}^{3 \times H \times W}$, a previous frame $\mathbf{I}_{t-1} \in \mathbb{R}^{3 \times H \times W}$, and a previous latent frame $\mathbf{H}_{t-1} \in \mathbb{R}^{2 \times H \times W}$.  Mathematically, \arch~takes the form:
\begin{equation}
    \mathbf{O}_t, \mathbf{H}_t = \mathcal{F}(\mathbf{I}_t, \mathbf{I}_{t-1}, \mathbf{H}_{t-1}) ,
\end{equation}
where $\mathcal{F}$ is our learned network, \arch, that efficiently synthesizes restored frame $\mathbf{O}_t$ and a latent frame $\mathbf{H}_t$, conditioned on inputs ($\mathbf{I}_t$, $\mathbf{I}_{t-1}$ and $\mathbf{H}_{t-1}$). Overall, \arch~has three main modules: (1) alignment module, (2) differential module, and (3) fusion module. 

\begin{figure}[t!]
    \centering
    \begin{subfigure}[b]{0.48\columnwidth}
         \centering
         \includegraphics[width=0.8\columnwidth]{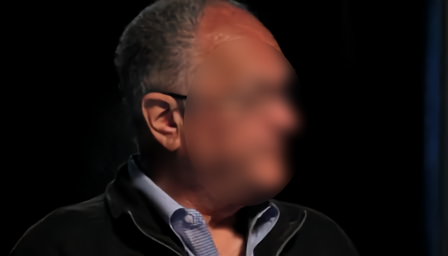}
         \caption{Previous frame}
     \end{subfigure}
     \hfill
     \begin{subfigure}[b]{0.48\columnwidth}
         \centering
         \includegraphics[width=0.8\columnwidth]{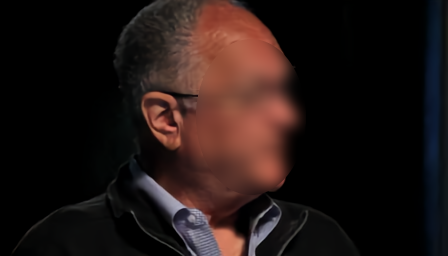}
         \caption{Current frame}
     \end{subfigure}
     \vfill
     \begin{subfigure}[b]{0.48\columnwidth}
         \centering
         \includegraphics[width=0.8\columnwidth]{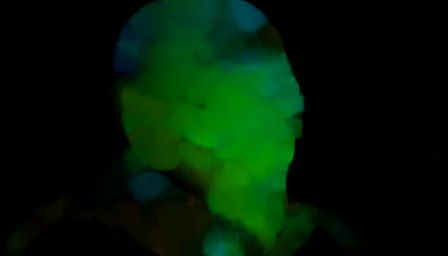}
         \caption{Optical flow}
         \label{fig:optical_flow}
     \end{subfigure}
     \hfill
     \begin{subfigure}[b]{0.48\columnwidth}
         \centering
         \includegraphics[width=0.8\columnwidth]{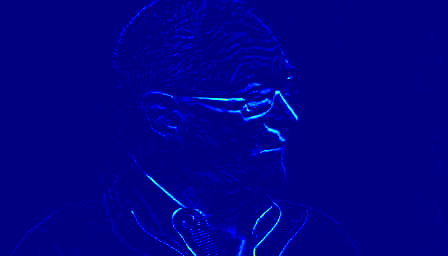}
         \caption{Difference (b - a)}
         \label{fig:difference_image}
     \end{subfigure}
     \vfill
     \begin{subfigure}[b]{0.48\columnwidth}
         \centering
         \includegraphics[width=0.8\columnwidth]{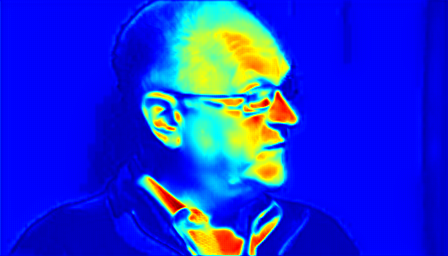}
         \caption{Alignment module output}
         \label{fig:alignment_output}
     \end{subfigure}
     \hfill
     \begin{subfigure}[b]{0.48\columnwidth}
         \centering
         \includegraphics[width=0.8\columnwidth]{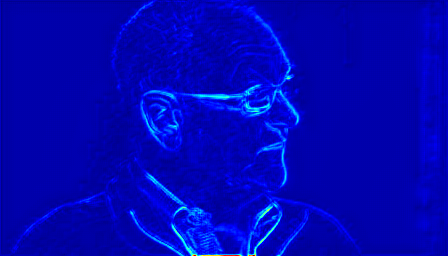}
         \caption{Differential module output}
         \label{fig:differential_output}
     \end{subfigure}
    \caption{This example visualizes outputs of two \arch~modules (alignment and  differential). The alignment module pays attention to areas corresponding to motion, i.e., nose and spectacles (c, d vs. e) while the differential module pays attention to high frequency components (e.g., spectacle edges in (f)) in region corresponding to motion.}
    \label{fig:mimic_optical_sharp}
\end{figure}

\vspace{1mm}
\noindent \textbf{Alignment module:} The alignment module takes a concatenation of the inputs ($\mathbf{I}_t$, $\mathbf{I}_{t-1}$ and $\mathbf{H}_{t-1}$) and produces aligned representations $\mathbf{A}_t \in \mathbb{R}^{d\times H \times W}$ using an efficient and light-weight encoder-decoder network (Figure \ref{fig:tiny_enc_dec}). The alignment module first learns pyramidal representations using the encoder network. These representations are then combined by the decoder to produce aligned representations. Compared to existing methods that learns very deep pyramidal representations for motion estimation \cite{lucas1981iterative, dosovitskiy2015flownet, IMKDB17, Sun_CVPR_2018, ranjan2017optical}, \arch~is very light-weight and shallow. To demonstrate the ability of \arch~in modeling the motion, an example is shown in Figure \ref{fig:mimic_optical_sharp} where person moves his head during a conversation. The most salient regions between consecutive frames are near the nose, spectacles, and shirt as depicted by the optical flow and difference image in Figure \ref{fig:optical_flow} and \ref{fig:difference_image}, respectively. The alignment module in the \arch~also pays attention to these salient regions (red color regions in Figure \ref{fig:differential_output}), illustrating that \arch~can model the motion implicitly. 

Specifically, the encoder in the alignment module consists of (a) a $5 \times 5$ standard convolutional layer, (b) a $5 \times 5$ standard convolutional layer with a stride of 2, (c) a point-wise convolutional layer, and (d) $N_A$ convolutional units (CUs; Section \ref{ssec:convolutional_unit}), where $N_A$ controls the depth of alignment module. The decoder follows a simplified UNet-like architecture \cite{ronneberger2015u}. The output of the last convolutional unit (CU) is first upsampled and then concatenated with the output of the first $5\times 5$ convolutional layer. The resultant output is then fused using a point-wise convolution to produce aligned representations $\mathbf{A}_t$. 

\vspace{1mm}
\noindent \textbf{Differential module:} The differential module aims at learning high-frequency components in an image such as object edges. To do so, the input $\mathbf{I}_t$ is first projected to the same dimensionality as $\mathbf{A}_t$ using a $3\times 3$ convolutional layer to produce a projected output $\mathbf{P}_t \in \mathbb{R}^{d\times H \times W}$. An element-wise difference is then computed between $\mathbf{P}_t$ and $\mathbf{A}_t$. The resultant output is then fed to differential module to further refine these representations and produce high-frequency representations $\mathbf{D}_t \in \mathbb{R}^{d\times H \times W}$. Figure \ref{fig:differential_output} shows an example where \arch~pays attention to high-frequency components (e.g., spectacle and ear edges). Similar to the alignment module, the differential module also takes the form of small and light-weight encoder-decoder network, with an exception to number of CUs. In the differential module, we stack $N_D$ CUs.

\vspace{1mm}
\noindent \textbf{Fusion module:} The fusion module combines high-frequency representations obtained from the differential module $\mathbf{D}_t$ with projected input representations $\mathbf{P}_t$ and produces restored frame $\mathbf{O}_t$ and latent frame $\mathbf{H}_t$. We first add $\mathbf{D}_t$ with $\mathbf{P}_t$ to enhance high-frequency components and then feed the resultant tensor to a fusion module. If the spatial dimensions of $\mathbf{O}_t$ are not the same as $\mathbf{I}_t$ (e.g., in super-resolution), the output of fusion module is up-sampled using a pixel-shuffle operation. Otherwise, an identity operation is performed. The resultant output is then convolved with a $3\times 3$ convolutional layer to produce $\mathbf{O}_t$. In parallel, the output of fusion layer is also projected using a point-wise convolutional layer to produce latent frame $\mathbf{H}_t$. Similar to the alignment and differential module, the fusion module is also an efficient and light-weight encoder-decoder network with $N_F$ CUs.

The operation of differential and fusion module is similar to traditional image enhancement methods (e.g., unsharp mask)  \cite{polesel2000image, deng2010generalized}. In such approaches, the input image is first smoothed to suppress high-frequency components. Then, a difference between smoothed image and input image is computed to identify high-frequency components, which are then added back to the input image to enhance it. 

\subsection{Convolutional Unit (CU)}
\label{ssec:convolutional_unit}
CNN-based methods for different visual recognition tasks learns representations using either a single branch (e.g., ResNet \cite{he2016deep} and MobileNets \cite{howard2017mobilenets, sandler2018mobilenetv2}) or multiple branches (e.g., InceptionNets \cite{szegedy2015going, szegedy2016inception} and ESPNets \cite{mehta2018espnet, mehta2019espnetv2}). We also study these two methods for learning representations. For learning representations at a single scale, we use a depth-wise convolutional layer with $7\times 7$ kernel while for learning representations at multiple scales, we apply three depth-wise convolutional layers simultaneously ($3\times 3$, $5\times 5$, and $7 \times 7$). In both of these methods, the effective receptive field is the same, i.e., $7 \times 7$. Following recent efficient architectures (e.g., MobileNetv3 \cite{howard2019searching}), we also adopt squeeze-excitation unit (SE unit) \cite{hu2018squeeze} to model channel inter-dependencies. Figure \ref{fig:conv_units} sketches the CU.

\begin{figure*}[t!]
    \centering
    \begin{subfigure}[b]{0.6\columnwidth}
         \centering
         \begin{subfigure}[b]{\columnwidth}
            \centering
            \includegraphics[height=85px]{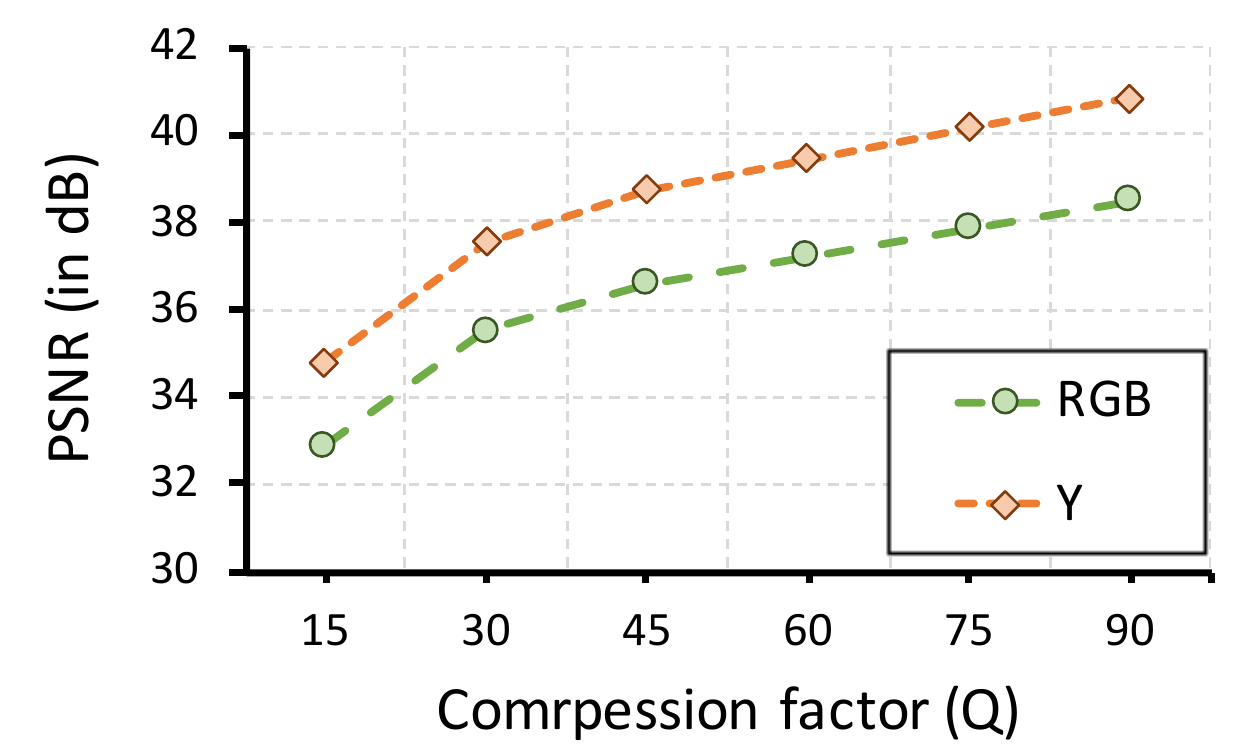}
            \caption{}
            \label{fig:compress_perf_psnr}
        \end{subfigure}
        \vfill
        \begin{subfigure}[b]{\columnwidth}
            \centering
            \includegraphics[height=85px]{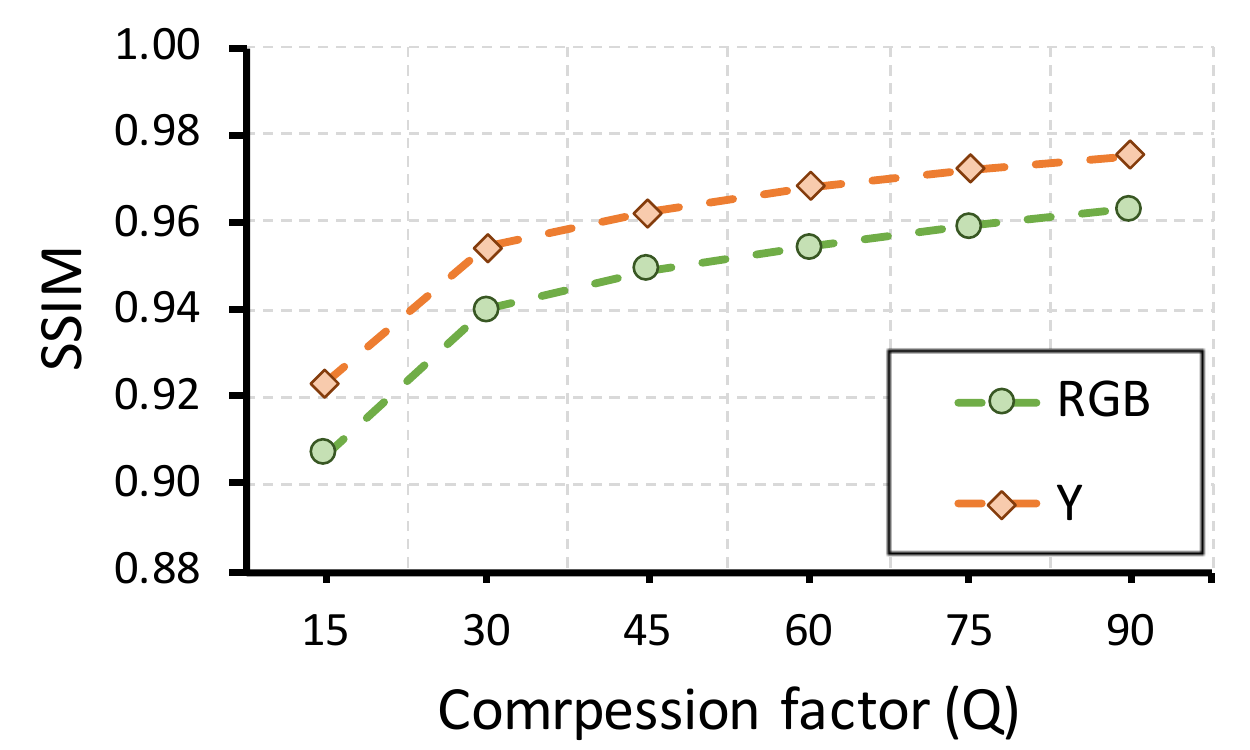}
            \caption{}
            \label{fig:compress_perf_ssim}
        \end{subfigure}
     \end{subfigure}
     \hfill
    \begin{subfigure}[b]{1.38\columnwidth}
         \centering
         \includegraphics[height=185px]{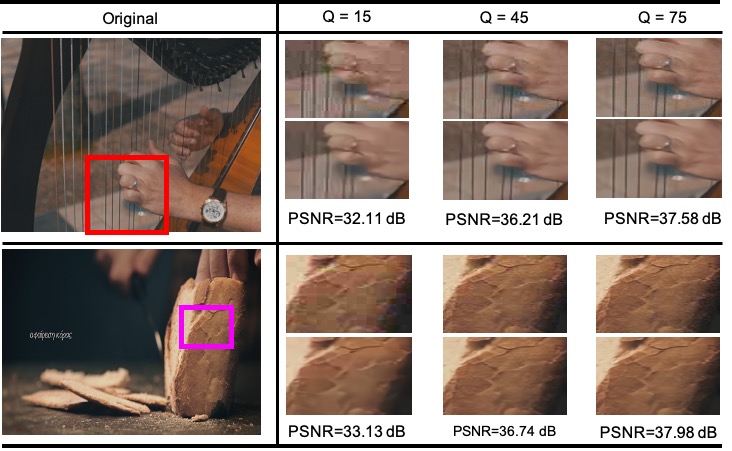}
         \caption{ }
         \label{fig:compress_example_qual}
     \end{subfigure}
    \caption{\textbf{Performance of \arch~under compression artifacts.} In \textbf{(a, b)}, performance in terms of PSNR and SSIM is measured as a function of compression factor $Q$ on both RGB and Y frames, respectively. Lower value of $Q$ means higher compression. In \textbf{(c)}, qualitative results for two sample images are shown at different value of $Q$. The top and bottom panels corresponds to the compressed frame and restored frames, respectively. Here, PSNR values are computed on RGB frames. See Appendix \ref{ssec:deblock_ablate} for more results.}
    \label{fig:compression_results}
\end{figure*}

\section{Experimental Results}
\label{sec:results}

To demonstrate the effectiveness of \arch~on video restoration tasks, we evaluate its performance on three tasks: (1) deblocking (Section \ref{ssec:video_deblock}), (2) denoising (Section \ref{ssec:video_denoise}), and (3) super-resolution (Section \ref{ssec:video_sr}). In this section, we first describe the experimental set-up and then evaluate the performance of \arch~on each of these tasks.

\subsection{Experimental Set-up}

\noindent \textbf{Tasks:} We study three video restoration tasks: (1) \textit{Video deblocking} aims at removing artifacts that may arise due to video compression, (2) \textit{Video denoising} aims at removing noise (e.g., adaptive white gaussian noise (AWGN)) which may be induced during video transmission, and (3) \textit{Video super-resolution} which aims at up-sampling low-resolution video to high-resolution at receiver's end.

\vspace{1mm}
\noindent \textbf{Dataset:} To evaluate the performance of \arch, we use large-scale Vimeo-90K dataset \cite{xue2019video} which consists of about 90K independent and diverse video shots with both indoor and outdoor lighting scenarios. We use official training and test splits. Note that, we split the training set randomly into two subsets using 90:10 ratio. The first subset is used for training while the second subset is used for validation. 

\vspace{1mm}
\noindent \textbf{Training:} \arch~models are trained by minimizing L1 loss using ADAM optimizer \cite{kingma2014adam} for 50 epochs (or about 50K iterations) using PyTorch. Based on our ablation experiments in Section \ref{sec:ablations}, we set $N_A=5$, $N_D=2$, and $N_F=2$.  The learning rate is increased linearly from $1e^{-7}$ to $1e^{-3}$ in first 100 iterations and is then annealed by half at 15-, 25-, 35-, and 45-th epochs. We train \arch~with an effective batch size of 64 (8 images per GPU x 8 GPUs) and use a L2 weight decay of $1e^{-6}$. All our convolutional layers are followed by a PReLU activation \cite{he2015delving}, except the activation in multi-scale block is after the addition operation. Standard augmentation methods, such as random crop, random flipping, random gamma correction, and random rotation, are used during training. Task-specific augmentation methods are included in respective sub-sections. For comparison with existing methods, we use official splits for deblocking, denoising, and super-resolution while for sensitivity studies, we use functions from OpenCV and Skimage libraries. 

\vspace{1mm}
\noindent \textbf{Evaluation metrics:} We use two standard quantitative metrics: (1) peak signal-to-noise ratio (PSNR) and (2) structural similarity index (SSIM). Higher value of PSNR and SSIM indicates better performance. Following previous methods, we report these metrics on RGB and Y channel (YCbCr color space).

\begin{figure*}[t!]
    \centering
    \begin{subfigure}[b]{2\columnwidth}
        \centering
        \begin{subfigure}[b]{0.245\columnwidth}
            \centering
            \includegraphics[width=\columnwidth]{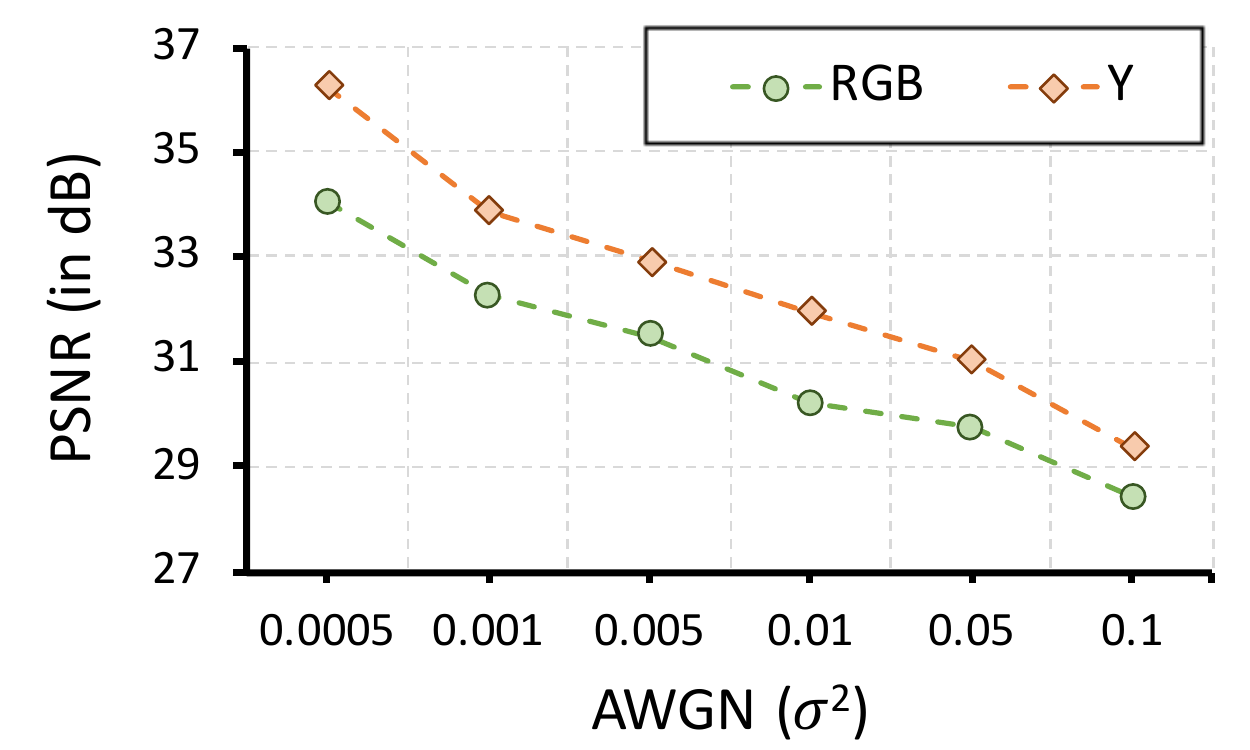}
            \caption{}
            \label{fig:awgn_psnr_perf}
        \end{subfigure}
        \hfill
        \begin{subfigure}[b]{0.245\columnwidth}
            \centering
            \includegraphics[width=\columnwidth]{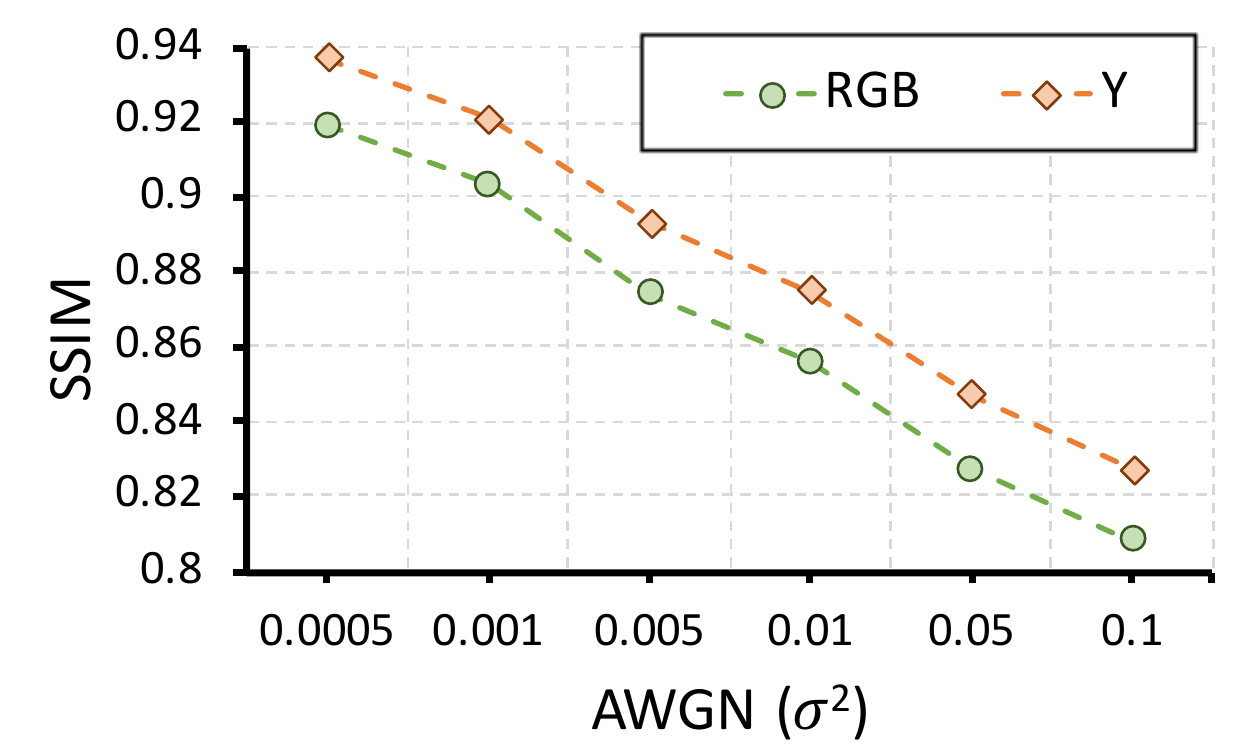}
            \caption{}
            \label{fig:awgn_ssim_perf}
        \end{subfigure}
        \hfill
        \begin{subfigure}[b]{0.245\columnwidth}
            \centering
            \includegraphics[width=\columnwidth]{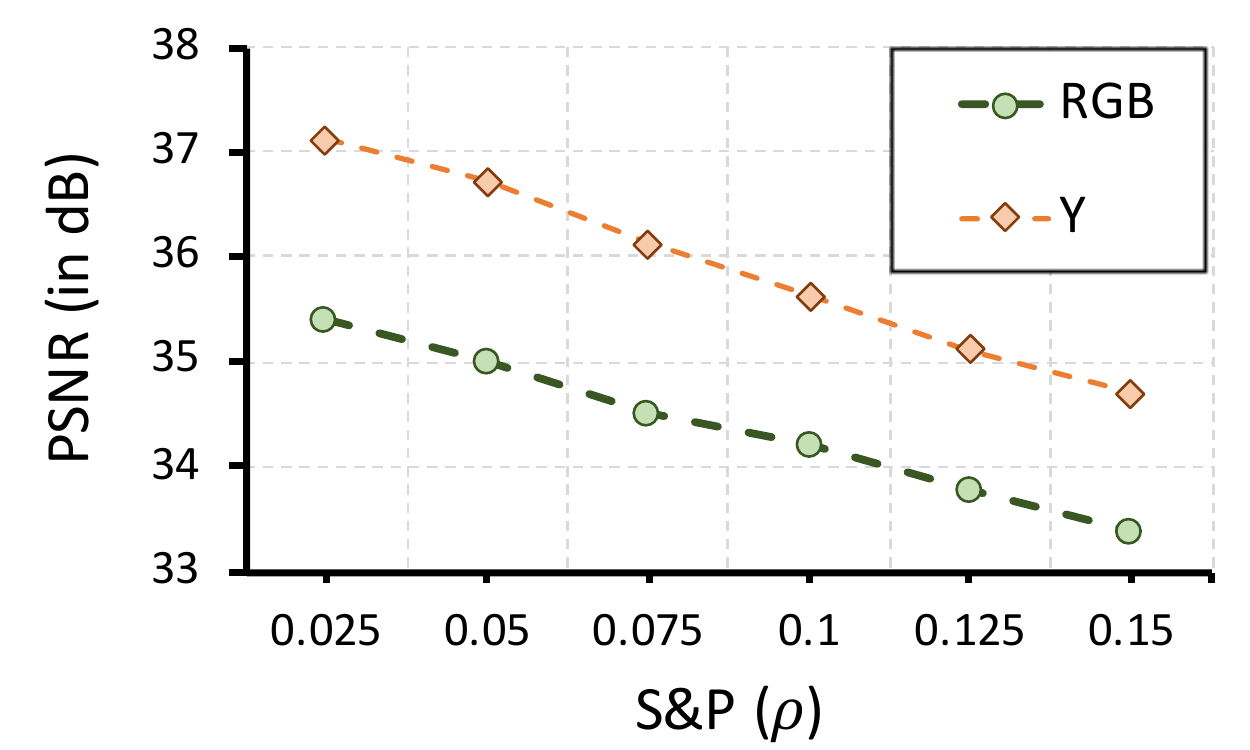}
            \caption{}
            \label{fig:snp_psnr_perf}
        \end{subfigure}
        \hfill
        \begin{subfigure}[b]{0.245\columnwidth}
            \centering
            \includegraphics[width=\columnwidth]{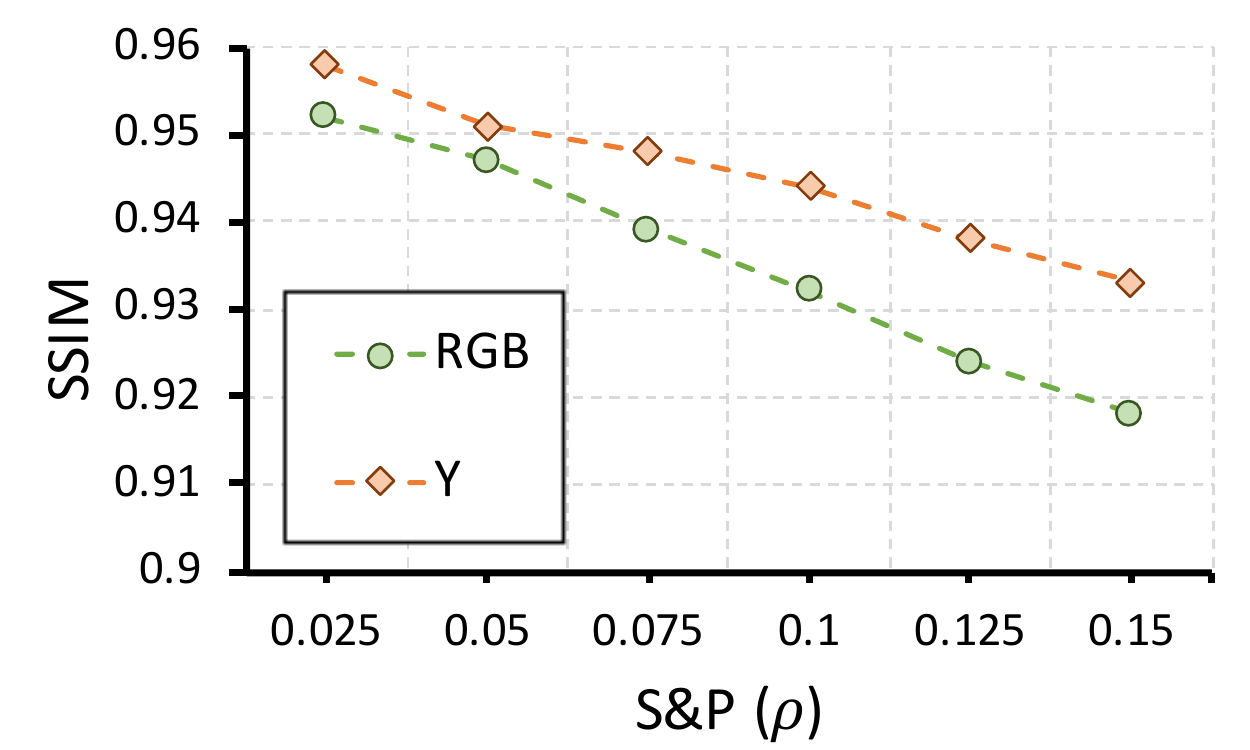}
            \caption{}
            \label{fig:snp_ssim_perf}
        \end{subfigure}
    \end{subfigure}
    \vfill
    \begin{subfigure}[b]{0.75\columnwidth}
         \centering
         \begin{subfigure}[b]{\columnwidth}
            \centering
            \includegraphics[height=80px]{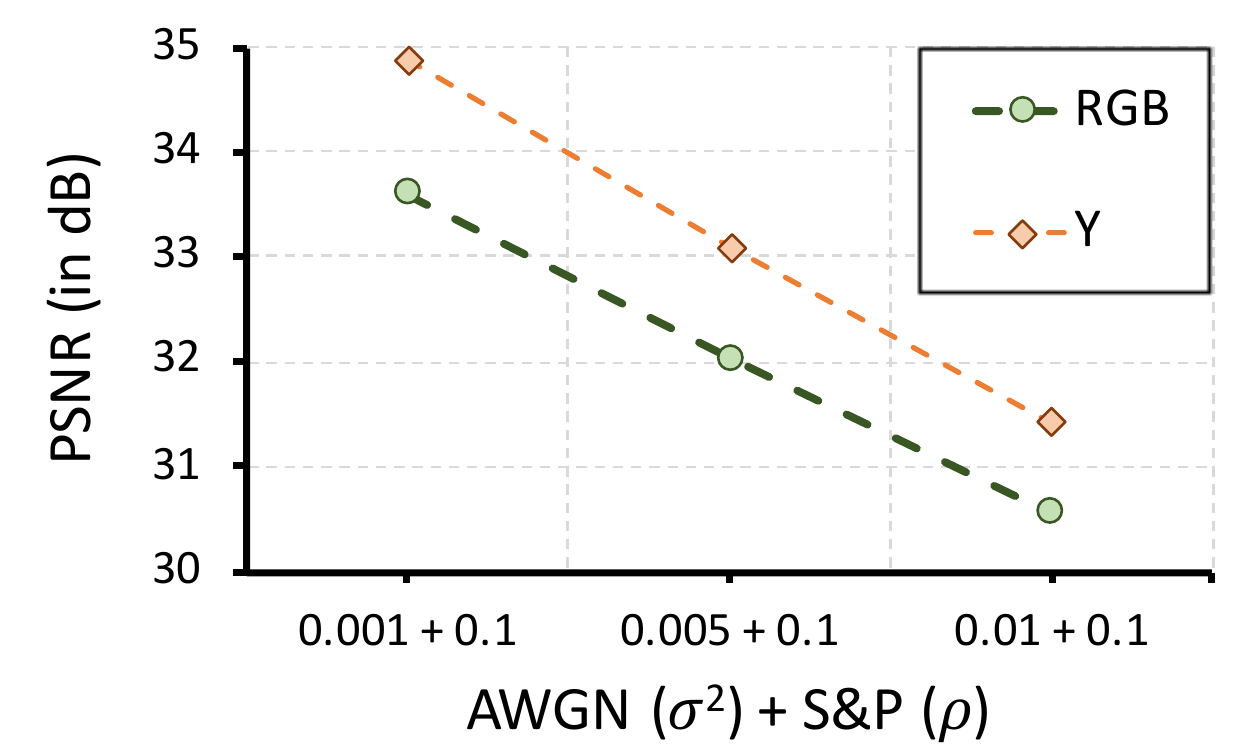}
            \caption{}
            \label{fig:mixed_psnr_perf}
        \end{subfigure}
        \vfill
        \begin{subfigure}[b]{\columnwidth}
            \centering
            \includegraphics[height=80px]{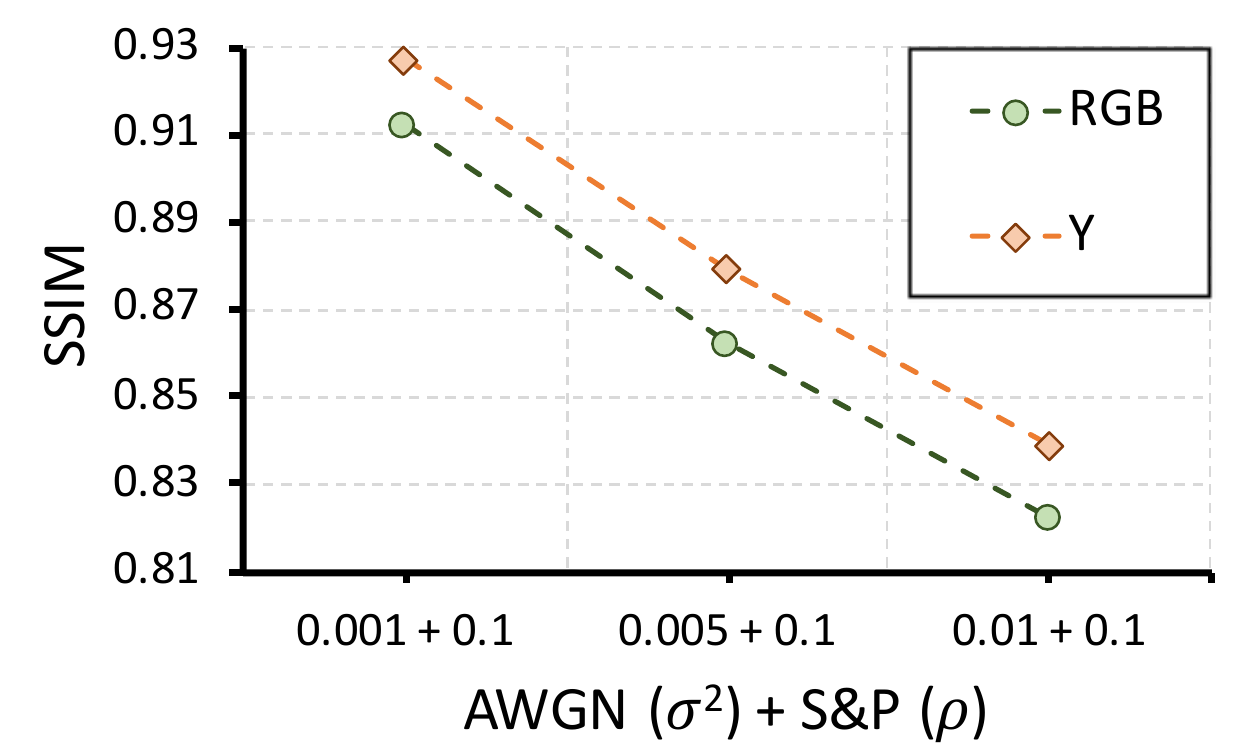}
            \caption{}
            \label{fig:mixed_ssim_perf}
        \end{subfigure}
     \end{subfigure}
     \hfill
    \begin{subfigure}[b]{1.2\columnwidth}
         \centering
         \hspace{-1.5cm}\includegraphics[height=180px]{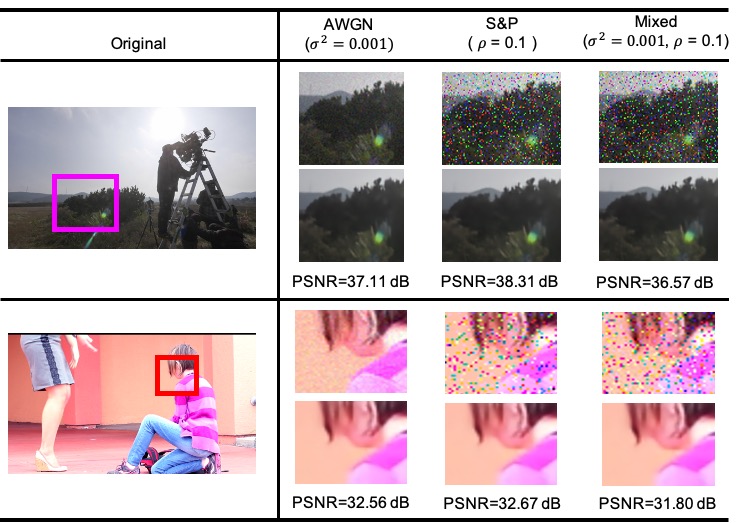}
         \caption{ }
         \label{fig:noise_qual_example}
     \end{subfigure}
    \caption{\textbf{Performance of \arch~under noise artifacts.} In \textbf{(a, b)}, performance in terms of PSNR and SSIM is measured as a function of AWGN noise varience $\sigma^2$ on both RGB and Y frames, respectively. Similarly, in \textbf{(c, d)} and \textbf{(e, f)}, performance curves are drawn for salt and pepper noise (S\&P) density $\rho$ and mixed noise (AWGN + S\&P). Lower value of $\sigma^2$ and $\rho$ means less noise. In \textbf{(g)}, qualitative results for two sample images are shown for different types of noise. The top and bottom panels corresponds to the noisy and restored frames, respectively. Here, PSNR values are for RGB frames. See Appendix \ref{ssec:denoise_ablate} for more results.}
    \label{fig:noise_robustness_study}
\end{figure*}

\subsection{Video Deblocking}
\label{ssec:video_deblock}

\noindent \textbf{Sensitivity study:} We train and evaluate \arch~for the task of deblocking artifacts. Similar to state-of-the-art methods (e.g., \cite{xue2019video, lu2018deep}), we compress frames using JPEG2000 compression. During training, we randomly select the compression or quality factor ($Q$) between 10 and 40. During evaluation, we vary the value of $Q$ from 15 to 90 using OpenCV. Smaller value of $Q$ indicates higher compression or more blocking artifacts. Note that the same \arch~network is evaluated at different values of $Q$.

Figure \ref{fig:compression_results} shows quantitative and qualitative results under different values of $Q$. The quantitative results in Figure \ref{fig:compress_perf_psnr} and Figure \ref{fig:compress_perf_ssim} for both RGB and Y-channel shows that \arch~is robust to compression. For example, at $Q=15$, \arch~is able to achieve PSNR and SSIM values (RGB space) of 33 dB and 0.91, respectively, indicating that it can generate good quality frames even under high compression. These quantitative results are further strengthened with the qualitative results in Figure \ref{fig:compress_example_qual}. The compression artifacts around the hand and strings of harp in the first row and bread loaf in the second row of Figure \ref{fig:compress_example_qual} are completely removed by \arch, even under high compression.

\vspace{1mm}
\noindent \textbf{Comparison with state-of-the-art methods:} Table \ref{tab:compress_sota} compares the performance of \arch~with state-of-the-art deblocking methods (ARCNN \cite{dong2015compression}, DnCNN \cite{zhang2017beyond}, V-BM4D \cite{maggioni2012video}, ToFlow \cite{xue2019video}, and DKFN \cite{lu2018deep}) on the official Vimeo-90K test set. \arch~delivers similar or better performance than existing methods while having significantly fewer network parameters and multiplication-addition operations (MACs). For example, \arch~delivers the similar performance as ToFlow \cite{xue2019video}, but has $46\times$ fewer MACs and $13.64\times$ fewer parameters.

\begin{table}[t!]
    \centering
    \resizebox{0.8\columnwidth}{!}{
    \begin{tabular}{lrrrr}
        \toprule[1.5pt]
        \textbf{Method} & \textbf{MACs} & \textbf{\# Params} & \textbf{PSNR} & \textbf{SSIM} \\
        \midrule[1pt]
        ARCNN$^\dagger$ \cite{dong2015compression} & 27.73 G & 117.73 K & 36.11 & 0.960 \\
        DnCNN$^\dagger$ \cite{zhang2017beyond} & 128.64 G & 558.34 K & 37.26 & 0.967 \\
        V-BM4D \cite{maggioni2012video} & -- & -- & 35.75 & 0.959 \\
        ToFlow \cite{xue2019video} & 466.83 G & 1073.48 K & 36.92 & 0.966 \\
        DKFN \cite{lu2018deep} & -- & -- & 37.93 & 0.971 \\
        \arch~(Ours) & 10.13 G & 78.71 K & 36.65 & 0.967 \\
        \bottomrule[1.5pt]
    \end{tabular}
    }
    \caption{\textbf{Comparison with state-of-the-art methods on the task of video deblocking.}. \arch~delivers similar or better performance, but with significantly fewer parameters and multiplication-addition operations (MACs). Here, we report results on the official Vimeo-90K compressed test set where frames are compressed using FFMPEG. See \cite{xue2019video} for more details. Similar to previous works, we report results in RGB colorspace. The results of methods marked with $^\dagger$ are reported in \cite{lu2018deep} while V-BM4D's performance is reported in \cite{xue2019video}. MACs are measured for $640\times360$ RGB frame.}
    \label{tab:compress_sota}
\end{table}

\subsection{Video Denoising}
\label{ssec:video_denoise}

\noindent \textbf{Sensitivity study:} Following state-of-the-art methods, we train and evaluate \arch~under three noise types: (1) Additive White Gaussian Noise (AWGN), (2) Salt and Pepper noise (S\&P), and (3) mixture of AWGN and S\&P. During training, we randomly select the variance of AWGN noise $\sigma^2$ between $0.05$ and $0.4$ and the density of S\&P noise $\rho$ between $0.05$ and $0.3$. Here, $\sigma$ represents the standard deviation and the value of $\rho$ measures the percentage of pixels randomly replaced with noise. For example, $\rho=0.3$ indicates that $30\%$ of pixels in a frame are randomly replaced with S\&P noise. During evaluation, we first study the effect of AWGN (Figure \ref{fig:awgn_psnr_perf} and \ref{fig:awgn_ssim_perf}) and S\&P (Figure \ref{fig:snp_psnr_perf} and \ref{fig:snp_ssim_perf}) independently. For AWGN, we vary $\sigma^2$ between $0.0005$ and $0.1$ while for S\&P, we vary $\rho$ between $0.025$ and $0.15$. We then study the effect of mixture of AWGN and S\&P noise (Figure \ref{fig:mixed_psnr_perf} and \ref{fig:mixed_ssim_perf}). In these experiments, we set $\rho=0.1$ and vary $\sigma^2$ between 0.001 and 0.1. Note that we train only one \arch~network for video denoising and then evaluate it at different settings of AWGN, S\&P, and mixed noise. 

The quantitative results in Figure \ref{fig:noise_robustness_study} shows that \arch~is robust to different types and amounts of noise. For example, the RGB PSNR values of \arch~with AWGN noise ($\sigma^2=0.001$; Figure \ref{fig:awgn_psnr_perf}), S\&P noise ($\rho=0.1$; Figure \ref{fig:snp_psnr_perf}), and mixed noise ($\sigma^2=0.001$ and $\rho=0.1$; Figure \ref{fig:mixed_psnr_perf}) are around 33 dB, showing the robustness of \arch~to different types of noise. This is further demonstrated qualitatively in Figure \ref{fig:noise_qual_example}. In the first and second row of Figure \ref{fig:noise_qual_example}, we can see that \arch~is able to remove noise and also, restore fine details (e.g., hairs in the second row) under different types of noise.
\begin{table}[t!]
    \centering
    \begin{subtable}[b]{\columnwidth}
        \centering
        \resizebox{0.8\columnwidth}{!}{
            \begin{tabular}{lrrrr}
            \toprule[1.5pt]
              \textbf{Method}  & \textbf{MACs} & \textbf{\# Params} & \textbf{PSNR} & \textbf{SSIM} \\
                \midrule[1.25pt]
                ToFlow \cite{xue2019video} & 466.83 G & 1073.48 K & 33.51 & 0.939 \\
                \arch~(Ours) & 10.13 G & 78.71 K & 32.37 & 0.921 \\
                \bottomrule[1.5pt]
            \end{tabular}
        }
        \caption{Vimeo-90K official test set}
        \label{tab:noise_vimeo}
    \end{subtable}
    \vfill
    \begin{subtable}[b]{\columnwidth}
        \centering
        \resizebox{0.7\columnwidth}{!}{
            \begin{tabular}{lrrr}
            \toprule[1.5pt]
              \textbf{Method}  & \textbf{MACs} & \textbf{\# Params} & \textbf{PSNR} \\
                \midrule[1.25pt]
                V-BM4D$^\dagger$ \cite{maggioni2012video} & -- & -- & 26.31 \\
                DnCNN$^\dagger$ \cite{zhang2017beyond} & 128.64 G & 588.34 K & 26.64 \\
                N2V $^{\star\dagger}$ \cite{krull2019noise2void} & 140.61 G & 27.90 M & 25.17 \\
                N2N+F2F \cite{yu2020joint} & -- & -- & 26.56 \\
                \arch~(Ours) & 10.13 G & 78.71 K & 25.79 \\
                \bottomrule[1.5pt]
            \end{tabular}
        }
        \caption{Vid4 dataset}
        \label{tab:noise_vid4}
    \end{subtable}
    \vfill
    \begin{subtable}[b]{\columnwidth}
        \centering
        \resizebox{\columnwidth}{!}{
        \begin{tabular}{cc}
            \includegraphics[width=0.6\columnwidth]{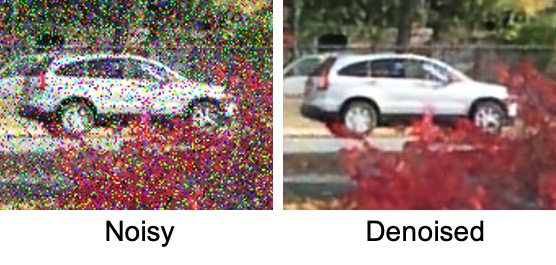} &  \includegraphics[width=0.385\columnwidth]{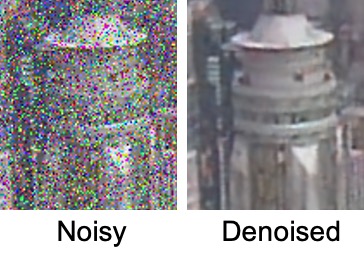}
        \end{tabular}
        }
        \caption{Qualitative denoising results using \arch~on Vid4 dataset.}
    \end{subtable}
    \caption{\textbf{Comparison with state-of-the-art methods on the task of video denoising.} \arch~is able to denoise videos efficiently. Similar to previous works, we report results in RGB colorspace. Here, $^\dagger$ represents results are from \cite{yu2020joint} and $^\star$ represents that the number of MACs and parameters are computed for U-Net \cite{ronneberger2015u} as N2V is built on top of U-Net. On Vid4 dataset, previous works have not reported SSIM. Therefore, we do not report SSIM on Vid4 dataset.}
    \label{tab:noise_results}
\end{table}

\vspace{1mm}
\noindent \textbf{Comparison with state-of-the-art methods:} Most state-of-the-art methods train denoising models on Vimeo-90K dataset and evaluate on Vid4 dataset \cite{liu2011bayesian}. Following these works, we adopt the same strategy and evaluate on Vid4 dataset. We also compare \arch~with ToFlow on the official Vimeo-90K denoising dataset. Results are shown in Table \ref{tab:noise_results}. \arch~delivers competitive performance to state-of-the-art methods, but with significantly fewer MACs and parameters. It is worth mentioning that some existing methods (e.g., ToFlow \cite{xue2019video} and N2N + F2F \cite{yu2020joint}) use optical flow, which is either computationally expensive or requires specialized accelerators. Unlike these methods, \arch~does not requires any flow estimation and is suitable for edge devices.

\subsection{Video Super-resolution}
\label{ssec:video_sr}
We train and evaluate \arch~on video super-resolution ($2 \times$ and $4 \times$) task. For training \arch~that upsamples the input by $2\times$, we randomly crop a patch whose size lies in the range: $\{128, 144, 160, 176, 192 \}$. For $4\times$ model, we finetune $2\times$ model and select random patch size in the range: $\{64, 72, 80, 88, 96\}$. Table \ref{tab:vimeo_super_res} shows that \arch~delivers competitive performance as compared to existing methods, but with significantly fewer parameters and MACs. For example, the SSIM score of \arch~is 0.018 lower than the EDVR, but has $260 \times$ and $958 \times$ fewer parameters and MACs, respectively.

\begin{table}[t!]
    \centering
    \resizebox{0.95\columnwidth}{!}{
        \begin{tabular}{lcrrrr}
        \toprule[1.5pt]
          \textbf{Method} & \textbf{Up-sampling}  & \textbf{MACs} & \textbf{\# Params} & \textbf{PSNR} & \textbf{SSIM} \\
            \midrule[1.25pt]
            ToFlow \cite{xue2019video} & $4\times$ & 466.83 G & 1073.48 K & 34.83 & 0.922 \\
            DUF \cite{jo2018deep} & $4\times$ & -- & -- & 36.37 & 0.939 \\
            RBPN \cite{haris2019recurrent} & $4\times$ & 29.62 T  & 12.77 M & 37.07 & 0.944 \\
            EDVR \cite{edvr2019wang} & $4\times$ & 9.96 T  & 20.10 M & 37.61 & 0.949 \\
            \arch~(Ours) & $4\times$ & 10.39 G & 79.55 K & 35.98 & 0.931 \\
            \arch~(Ours) & $2\times$ & 10.13 G & 78.71 K & 37.86 & 0.965 \\
            \bottomrule[1.5pt]
        \end{tabular}
    }
    \caption{\textbf{Comparison with state-of-the-art methods on the task of super-resolution.} \arch~delivers competitive performance to existing methods, but with significantly fewer multiplication-addition operations (MACs) and network parameters. Similar to previous works, we report results in Y-channel on the official Vimeo-90K test set. See Appendix \ref{ssec:vsr_ablate} for qualitative results.}
    \label{tab:vimeo_super_res}
\end{table}

\section{Discussion}
\label{sec:discussion}
\noindent \textbf{Generalization to unseen dataset:} A video transmission system, shown in Figure \ref{fig:video_transmission_system}, compresses the video stream before transmitting to the destination in order to reduce network bandwidth. At the destination, the decoded video stream is of low quality due to compression and transmission noise, and is restored using the video restoration methods. 
To demonstrate the effectiveness of \arch~in real-world applications (e.g., real-time video conferencing), we trained ``multi-task" \arch~model that is capable of denoising and deblocking on edge devices (see Figure \ref{fig:video_transmission_system}). To train this model, we used the same training and validation sets as mentioned in Section \ref{sec:results}, with an exception to inputs to the model. During training, the input sequences were  randomly compressed ($Q \in \left[10, 40\right]$). After that, mixed noise ($\sigma^2 \in \left[0.001, 0.01\right]$ and $\rho \in \left[0.025, 0.15\right]$) is added to synthesize transmission noise. Each sequence in Vimeo-90K dataset comprises of 8 frames, has a fixed spatial resolution of $448 \times 256$, and are compressed frame-by-frame. Therefore, to test the ability of \arch~in modeling variable-length sequences under both camera and object motion, we evaluated its performance on six high-definition and diverse video sequences that are captured using different mobile devices (see Table \ref{tab:vid_details_unseen}). For evaluation, we first compressed these videos using H264 encoding and then added a mixed noise (AWGN with $\sigma^2=0.001$ and S\&P with $\rho=0.1$). Quantitative (Table \ref{tab:vid_details_unseen}) and qualitative (Figure \ref{fig:taser_examples}) results shows that \arch~(1) can model variable-length sequences and (2) generalizes to unseen videos.

\begin{table}[]
\resizebox{\columnwidth}{!}{
\begin{tabular}{lccclllcll}
    \toprule[1.5pt]
        &           & \multicolumn{2}{c}{\bfseries File Size} &  & \multicolumn{2}{c}{\bfseries RGB} && \multicolumn{2}{c}{\bfseries  Y-Channel} \\
        \cmidrule[1.25pt]{3-4} \cmidrule[1.25pt]{6-7} \cmidrule[1.25pt]{9-10}
\textbf{Seq. Id} & \textbf{\# Frames} & \textbf{Original} & \textbf{Compressed} &  & \textbf{PSNR} & \textbf{SSIM} && \textbf{PSNR} & \textbf{SSIM}\\
\midrule[1pt]
Seq-1   & 200       & 10.7 MB      & 1.43 MB        &  & 37.930      & 0.966     && 39.405         & 0.973        \\
Seq-2   & 200       & 35.54 MB     & 4.60 MB        &  & 35.662      & 0.963     && 36.730         & 0.971        \\
Seq-3   & 200       & 36.07 MB     & 4.74 MB        &  & 35.880      & 0.962     && 36.713         & 0.967        \\
Seq-4   & 915       & 56.66 MB     & 9.28 MB        &  & 38.320      & 0.976     && 39.656         & 0.981        \\
Seq-5   & 366       & 11.40 MB     & 8.05 MB        &  & 40.386      & 0.978     && 42.536         & 0.984        \\
Seq-6   & 821       & 10.57 MB     & 7.24 MB        &  & 38.775      & 0.974     && 40.903         & 0.979        \\
\midrule
\multicolumn{4}{l}{\bfseries Avg.}                            && 37.826      & 0.970     && 39.324         & 0.976   \\
\bottomrule[1.5pt]
\end{tabular}
}
\caption{\textbf{Quantitative results on unseen videos.} For generating videos with artifacts, videos are first compressed using H264 compression method. A mixed noise (AWGN with $\sigma^2=0.001$ and S\&P with $\rho=0.1$) is then added to synthesize transmission noise. \arch~is able to remove these artifacts, as is evident in Figure \ref{fig:taser_examples}.}
\label{tab:vid_details_unseen}
\end{table}

\begin{table}[t!]
    \centering
    \resizebox{0.8\columnwidth}{!}{
    \begin{tabular}{lcclcclcc}
        \toprule[1.5pt]
        \textbf{Input size}  & \multicolumn{2}{c}{\bfseries 240p} && \multicolumn{2}{c}{\bfseries 360p} && \multicolumn{2}{c}{\bfseries 480p} \\
        \cmidrule[1.25pt]{2-3} \cmidrule[1.25pt]{5-6} \cmidrule[1.25pt]{8-9}
        \textbf{Output size} & \textbf{240p} & \textbf{480p}   && \textbf{360p} & \textbf{720p} && \textbf{480p} & \textbf{960p}  \\
        \midrule[1pt]
        \textbf{iPhone XS} & 12.7 & 12.8 && 7.2 & 7.8 && 4.2 & 4.2 \\
        \textbf{iPhone 11} & 20.6 & 20.4 && 9.2 & 9.1 && 5.6 & 5.7 \\
        \bottomrule[1.5pt]
    \end{tabular}
}
\caption{\textbf{\arch's speed (in FPS) on edge devices.} Each data point is an average across 100 iterations.}
\label{tab:speed_iphone}
\end{table}

\vspace{1mm}
\noindent \textbf{Run-time on edge device:} Typically, video conference applications on edge devices processes 240p and 360p videos at 10-15 frames per second (FPS). To demonstrate the applicability of \arch~on edge devices, we measured it's inference time on two iOS devices: (1) iPhone XS and (2) iPhone 11. Table \ref{tab:speed_iphone} shows that \arch~runs in real-time. We would like to highlight that CoreML (Apple's ML engine) does not support PixelShuffle on the accelerator. To do that operation, we used a solution that uses reshape and transpose operations. These operations are performed on iPhone's CPU (23\% CPU occupancy), which resulted in drop in speed. Also, \arch~is faster on iPhone 11 in comparison to iPhone XS. We believe that accelerator-specific implementations of PixelShuffle along with advancements in hardware technology would further improve the speed of \arch~on edge devices.

\begin{table}[t!]
\centering
\begin{subtable}[b]{\columnwidth}
    \centering
    \resizebox{0.9\columnwidth}{!}{
    \begin{tabular}{lcrrrrcrr}
    \toprule[1.5pt]
            &         &         &            & \multicolumn{2}{c}{\bfseries RGB} && \multicolumn{2}{c}{\bfseries Y-Channel} \\
            \cmidrule[1pt]{5-6}\cmidrule[1pt]{8-9}
    {\bfseries CU Type} & {\bfseries SE Unit} & {\bfseries MACs}    & {\bfseries \# Params} & {\bfseries PSNR}        & {\bfseries SSIM}      && {\bfseries PSNR} & {\bfseries SSIM}         \\
    \midrule[1pt]
    Single  & \xmark       & 9.85 G  & 68.15 K    & 31.207      & 0.868     && 32.650         & 0.886        \\
    Single  & \cmark       & 9.85 G  & 72.95 K    & 32.006      & 0.896     && 33.365         & 0.914        \\
    \midrule
    Multi   & \xmark       & 10.79 G & 73.91 K    & 29.026      & 0.875     && 30.247         & 0.895        \\
    \rowcolor{gray!30}
    Multi   & \cmark       & 10.79 G & 78.71 K    & 32.370      & 0.900     && 33.679         & 0.916       \\
    \bottomrule[1.5pt]
    \end{tabular}
    }
    \caption{Effect of different CU units}
    \label{tab:ablate_cu_units}
\end{subtable}
\vfill
\begin{subtable}[b]{\columnwidth}
    \centering
    \resizebox{0.9\columnwidth}{!}{
    \begin{tabular}{cccrrrrcrr}
    \toprule[1.5pt]
    \multicolumn{3}{c}{\bfseries Module depth} & & & \multicolumn{2}{c}{\bfseries RGB} && \multicolumn{2}{c}{\bfseries Y-Channel} \\
    \cmidrule[1pt]{1-3}\cmidrule[1pt]{6-7}\cmidrule[1pt]{9-10}
    $N_A$ & $N_D$ & $N_F$ & {\bfseries MACs}    & {\bfseries \# Params} & {\bfseries PSNR}        & {\bfseries SSIM}      && {\bfseries PSNR} & {\bfseries SSIM}  \\
    \midrule[1.25pt]
    1 & 1 & 7 & 11.44 G & 78.71 K & 31.605 & 0.887 && 32.913 & 0.905 \\
1 & 7 & 1 & 11.44 G & 78.71 K & 31.753 & 0.884 && 32.951 & 0.901 \\
7 & 1 & 1 & 9.47 G  & 78.71 K & 30.859 & 0.871 && 32.139 & 0.890 \\
\midrule
2 & 2 & 5 & 11.11 G & 78.71 K & 32.139 & 0.901 && 33.477 & 0.919 \\
2 & 5 & 2 & 11.11 G & 78.71 K & 32.057 & 0.891 && 33.445 & 0.908 \\
\rowcolor{gray!30}
5 & 2 & 2 & 10.13 G & 78.71 K & 32.403 & 0.903 && 33.884 & 0.921 \\
\midrule
3 & 2 & 4 & 10.77 G & 78.71 K & 31.690 & 0.890 && 33.047 & 0.908 \\
3 & 4 & 2 & 10.77 G & 78.71 K & 30.785 & 0.874 && 32.193 & 0.896 \\
4 & 3 & 2 & 10.46 G & 78.71 K & 31.416 & 0.877 && 32.690 & 0.895 \\
\midrule
3 & 3 & 3 & 10.79 G & 78.71 K & 32.370 & 0.900 && 33.679 & 0.916 \\
    \bottomrule[1.5pt]
    \end{tabular}
    }
    \caption{Effect of depth of alignment, differential, and fusion modules}
    \label{tab:ablate_depth}
\end{subtable}
\caption{\textbf{Ablation studies on the task of AWGN denoising ($\sigma^2 = 0.001$)}. Overall, \arch~with multi-scale CUs + SE unit and deeper alignment modules provides the best trade-off between performance and MACs.}
\label{tab:ablations}
\end{table}

\section{Ablations}
\label{sec:ablations}

\noindent \textbf{Effect of different CUs:}  Table \ref{tab:ablate_cu_units} studies the effect of single- and multi-scale convolutional units (CUs) with and without SE unit on the task of AWGN denoising. Multi-scale CU units with SE improves the performance. We hypothesize that this is because AWGN noise is identically distributed in the frames and kernels at different scales helps learn better representations and remove noisy artifacts (see gray color row in Table \ref{tab:ablate_cu_units}).

\vspace{1mm}
\noindent \textbf{Effect of the depth of alignment, differential, and fusion modules:} Table \ref{tab:ablate_depth} studies \arch~with different values of $N_A$, $N_D$, and $N_F$. We are interested in efficient networks for edge devices, therefore, we studied only those combinations that satisfies this criteria: $N_A + N_D + N_F=9$. We found that deeper alignment modules delivers the best trade-off between performance and MACs. Therefore, in our main experiments, we used $N_A=5$, $N_D=2$, and $N_F=2$ (see gray color row in Table \ref{tab:ablate_depth}). 

We perform similar studies for deblocking and super-resolution tasks (see Appendix \ref{sec:ablations_append}). We do not observe much gains with different CUs as well as varying the depth of alignment, differential, and fusion modules.

\section{Conclusion}
With virtual presence becoming more and more prominent in present days, it is imperative that video quality is perceptually pleasing which in turn makes the user's experience pleasant. This work introduces \arch, a unified network for deblocking, denoising, and super-resolving frames on edge devices. Compared to state-of-the-art video restoration models, \arch~is more efficient and runs in real-time on edge devices while delivering competitive performance across different tasks.  We believe this work will open up new research directions in this area. 

{\small
\bibliographystyle{ieee_fullname}
\bibliography{esrnet}

\begin{thebibliography}{10}\itemsep=-1pt

\bibitem{andri2018yodann}
Renzo Andri, Lukas Cavigelli, Davide Rossi, and Luca Benini.
\newblock Yodann: An architecture for ultralow power binary-weight cnn
  acceleration.
\newblock {\em IEEE Transactions on Computer-Aided Design of Integrated
  Circuits and Systems}, 2018.

\bibitem{bao2019memc}
Wenbo Bao, Wei-Sheng Lai, Xiaoyun Zhang, Zhiyong Gao, and Ming-Hsuan Yang.
\newblock Memc-net: Motion estimation and motion compensation driven neural
  network for video interpolation and enhancement.
\newblock {\em IEEE transactions on pattern analysis and machine intelligence},
  2019.

\bibitem{caballero2017real}
Jose Caballero, Christian Ledig, Andrew Aitken, Alejandro Acosta, Johannes
  Totz, Zehan Wang, and Wenzhe Shi.
\newblock Real-time video super-resolution with spatio-temporal networks and
  motion compensation.
\newblock In {\em Proceedings of the IEEE Conference on Computer Vision and
  Pattern Recognition}, pages 4778--4787, 2017.

\bibitem{chollet2017xception}
Fran{\c{c}}ois Chollet.
\newblock Xception: Deep learning with depthwise separable convolutions.
\newblock In {\em Proceedings of the IEEE conference on computer vision and
  pattern recognition}, pages 1251--1258, 2017.

\bibitem{deng2010generalized}
Guang Deng.
\newblock A generalized unsharp masking algorithm.
\newblock {\em IEEE transactions on Image Processing}, 20(5):1249--1261, 2010.

\bibitem{dong2015compression}
Chao Dong, Yubin Deng, Chen Change~Loy, and Xiaoou Tang.
\newblock Compression artifacts reduction by a deep convolutional network.
\newblock In {\em Proceedings of the IEEE International Conference on Computer
  Vision}, pages 576--584, 2015.

\bibitem{dong2014learning}
Chao Dong, Chen~Change Loy, Kaiming He, and Xiaoou Tang.
\newblock Learning a deep convolutional network for image super-resolution.
\newblock In {\em European conference on computer vision}, pages 184--199.
  Springer, 2014.

\bibitem{dosovitskiy2015flownet}
Alexey Dosovitskiy, Philipp Fischer, Eddy Ilg, Philip Hausser, Caner Hazirbas,
  Vladimir Golkov, Patrick Van Der~Smagt, Daniel Cremers, and Thomas Brox.
\newblock Flownet: Learning optical flow with convolutional networks.
\newblock In {\em Proceedings of the IEEE international conference on computer
  vision}, pages 2758--2766, 2015.

\bibitem{gupta2016cross}
Saurabh Gupta, Judy Hoffman, and Jitendra Malik.
\newblock Cross modal distillation for supervision transfer.
\newblock In {\em CVPR}, pages 2827--2836, 2016.

\bibitem{han2015deep}
Song Han, Huizi Mao, and William~J Dally.
\newblock Deep compression: Compressing deep neural networks with pruning,
  trained quantization and huffman coding.
\newblock In {\em ICLR}, 2016.

\bibitem{haris2019recurrent}
Muhammad Haris, Gregory Shakhnarovich, and Norimichi Ukita.
\newblock Recurrent back-projection network for video super-resolution.
\newblock In {\em Proceedings of the IEEE Conference on Computer Vision and
  Pattern Recognition}, pages 3897--3906, 2019.

\bibitem{he2015delving}
Kaiming He, Xiangyu Zhang, Shaoqing Ren, and Jian Sun.
\newblock Delving deep into rectifiers: Surpassing human-level performance on
  imagenet classification.
\newblock In {\em Proceedings of the IEEE international conference on computer
  vision}, pages 1026--1034, 2015.

\bibitem{he2016deep}
Kaiming He, Xiangyu Zhang, Shaoqing Ren, and Jian Sun.
\newblock Deep residual learning for image recognition.
\newblock In {\em Proceedings of the IEEE conference on computer vision and
  pattern recognition}, pages 770--778, 2016.

\bibitem{he2018amc}
Yihui He, Ji Lin, Zhijian Liu, Hanrui Wang, Li-Jia Li, and Song Han.
\newblock Amc: Automl for model compression and acceleration on mobile devices.
\newblock In {\em ECCV}, 2018.

\bibitem{hinton2015distilling}
Geoffrey Hinton, Oriol Vinyals, and Jeff Dean.
\newblock Distilling the knowledge in a neural network.
\newblock In {\em NIPS Deep Learning and Representation Learning Workshop},
  2015.

\bibitem{howard2019searching}
Andrew Howard, Mark Sandler, Grace Chu, Liang-Chieh Chen, Bo Chen, Mingxing
  Tan, Weijun Wang, Yukun Zhu, Ruoming Pang, Vijay Vasudevan, et~al.
\newblock Searching for mobilenetv3.
\newblock In {\em Proceedings of the IEEE International Conference on Computer
  Vision}, pages 1314--1324, 2019.

\bibitem{howard2017mobilenets}
Andrew~G Howard, Menglong Zhu, Bo Chen, Dmitry Kalenichenko, Weijun Wang,
  Tobias Weyand, Marco Andreetto, and Hartwig Adam.
\newblock Mobilenets: Efficient convolutional neural networks for mobile vision
  applications.
\newblock {\em arXiv preprint arXiv:1704.04861}, 2017.

\bibitem{hu2018squeeze}
Jie Hu, Li Shen, and Gang Sun.
\newblock Squeeze-and-excitation networks.
\newblock In {\em Proceedings of the IEEE conference on computer vision and
  pattern recognition}, pages 7132--7141, 2018.

\bibitem{huang2015bidirectional}
Yan Huang, Wei Wang, and Liang Wang.
\newblock Bidirectional recurrent convolutional networks for multi-frame
  super-resolution.
\newblock In {\em Advances in Neural Information Processing Systems}, pages
  235--243, 2015.

\bibitem{courbariaux2016binarized}
Itay Hubara, Matthieu Courbariaux, Daniel Soudry, Ran El-Yaniv, and Yoshua
  Bengio.
\newblock Binarized neural networks.
\newblock In {\em NIPS}, 2016.

\bibitem{IMKDB17}
E. Ilg, N. Mayer, T. Saikia, M. Keuper, A. Dosovitskiy, and T. Brox.
\newblock Flownet 2.0: Evolution of optical flow estimation with deep networks.
\newblock In {\em IEEE Conference on Computer Vision and Pattern Recognition
  (CVPR)}, Jul 2017.

\bibitem{jo2018deep}
Younghyun Jo, Seoung Wug~Oh, Jaeyeon Kang, and Seon Joo~Kim.
\newblock Deep video super-resolution network using dynamic upsampling filters
  without explicit motion compensation.
\newblock In {\em Proceedings of the IEEE conference on computer vision and
  pattern recognition}, pages 3224--3232, 2018.

\bibitem{kim2016accurate}
Jiwon Kim, Jung Kwon~Lee, and Kyoung Mu~Lee.
\newblock Accurate image super-resolution using very deep convolutional
  networks.
\newblock In {\em Proceedings of the IEEE conference on computer vision and
  pattern recognition}, pages 1646--1654, 2016.

\bibitem{kingma2014adam}
Diederik~P Kingma and Jimmy Ba.
\newblock Adam: A method for stochastic optimization.
\newblock {\em arXiv preprint arXiv:1412.6980}, 2014.

\bibitem{krull2019noise2void}
Alexander Krull, Tim-Oliver Buchholz, and Florian Jug.
\newblock Noise2void-learning denoising from single noisy images.
\newblock In {\em Proceedings of the IEEE Conference on Computer Vision and
  Pattern Recognition}, pages 2129--2137, 2019.

\bibitem{ledig2017photo}
Christian Ledig, Lucas Theis, Ferenc Husz{\'a}r, Jose Caballero, Andrew
  Cunningham, Alejandro Acosta, Andrew Aitken, Alykhan Tejani, Johannes Totz,
  Zehan Wang, et~al.
\newblock Photo-realistic single image super-resolution using a generative
  adversarial network.
\newblock In {\em Proceedings of the IEEE conference on computer vision and
  pattern recognition}, pages 4681--4690, 2017.

\bibitem{li2018constrained}
Chong Li and CJ~Richard Shi.
\newblock Constrained optimization based low-rank approximation of deep neural
  networks.
\newblock In {\em ECCV}, 2018.

\bibitem{liu2011bayesian}
Ce Liu and Deqing Sun.
\newblock A bayesian approach to adaptive video super resolution.
\newblock In {\em CVPR 2011}, pages 209--216. IEEE, 2011.

\bibitem{liu2017robust}
Ding Liu, Zhaowen Wang, Yuchen Fan, Xianming Liu, Zhangyang Wang, Shiyu Chang,
  and Thomas Huang.
\newblock Robust video super-resolution with learned temporal dynamics.
\newblock In {\em Proceedings of the IEEE International Conference on Computer
  Vision}, pages 2507--2515, 2017.

\bibitem{lu2018deep}
Guo Lu, Wanli Ouyang, Dong Xu, Xiaoyun Zhang, Zhiyong Gao, and Ming-Ting Sun.
\newblock Deep kalman filtering network for video compression artifact
  reduction.
\newblock In {\em Proceedings of the European Conference on Computer Vision
  (ECCV)}, pages 568--584, 2018.

\bibitem{lucas1981iterative}
Bruce~D Lucas, Takeo Kanade, et~al.
\newblock An iterative image registration technique with an application to
  stereo vision.
\newblock 1981.

\bibitem{ma2018shufflenet}
Ningning Ma, Xiangyu Zhang, Hai-Tao Zheng, and Jian Sun.
\newblock Shufflenet v2: Practical guidelines for efficient cnn architecture
  design.
\newblock In {\em Proceedings of the European conference on computer vision
  (ECCV)}, pages 116--131, 2018.

\bibitem{maggioni2012video}
Matteo Maggioni, Giacomo Boracchi, Alessandro Foi, and Karen Egiazarian.
\newblock Video denoising, deblocking, and enhancement through separable 4-d
  nonlocal spatiotemporal transforms.
\newblock {\em IEEE Transactions on image processing}, 21(9):3952--3966, 2012.

\bibitem{mehta2018espnet}
Sachin Mehta, Mohammad Rastegari, Anat Caspi, Linda Shapiro, and Hannaneh
  Hajishirzi.
\newblock Espnet: Efficient spatial pyramid of dilated convolutions for
  semantic segmentation.
\newblock In {\em Proceedings of the european conference on computer vision
  (ECCV)}, pages 552--568, 2018.

\bibitem{mehta2019espnetv2}
Sachin Mehta, Mohammad Rastegari, Linda Shapiro, and Hannaneh Hajishirzi.
\newblock Espnetv2: A light-weight, power efficient, and general purpose
  convolutional neural network.
\newblock In {\em Proceedings of the IEEE conference on computer vision and
  pattern recognition}, pages 9190--9200, 2019.

\bibitem{molchanov2019importance}
Pavlo Molchanov, Arun Mallya, Stephen Tyree, Iuri Frosio, and Jan Kautz.
\newblock Importance estimation for neural network pruning.
\newblock In {\em Proceedings of the IEEE Conference on Computer Vision and
  Pattern Recognition}, pages 11264--11272, 2019.

\bibitem{polesel2000image}
Andrea Polesel, Giovanni Ramponi, and V~John Mathews.
\newblock Image enhancement via adaptive unsharp masking.
\newblock {\em IEEE transactions on image processing}, 9(3):505--510, 2000.

\bibitem{ranjan2017optical}
Anurag Ranjan and Michael~J Black.
\newblock Optical flow estimation using a spatial pyramid network.
\newblock In {\em Proceedings of the IEEE Conference on Computer Vision and
  Pattern Recognition}, pages 4161--4170, 2017.

\bibitem{rastegari2016xnor}
Mohammad Rastegari, Vicente Ordonez, Joseph Redmon, and Ali Farhadi.
\newblock {Xnor-net: Imagenet classification using binary convolutional neural
  networks}.
\newblock In {\em ECCV}, 2016.

\bibitem{ronneberger2015u}
Olaf Ronneberger, Philipp Fischer, and Thomas Brox.
\newblock U-net: Convolutional networks for biomedical image segmentation.
\newblock In {\em International Conference on Medical image computing and
  computer-assisted intervention}, pages 234--241. Springer, 2015.

\bibitem{sajjadi2018frame}
Mehdi~SM Sajjadi, Raviteja Vemulapalli, and Matthew Brown.
\newblock Frame-recurrent video super-resolution.
\newblock In {\em Proceedings of the IEEE Conference on Computer Vision and
  Pattern Recognition}, pages 6626--6634, 2018.

\bibitem{sandler2018mobilenetv2}
Mark Sandler, Andrew Howard, Menglong Zhu, Andrey Zhmoginov, and Liang-Chieh
  Chen.
\newblock Mobilenetv2: Inverted residuals and linear bottlenecks.
\newblock In {\em Proceedings of the IEEE conference on computer vision and
  pattern recognition}, pages 4510--4520, 2018.

\bibitem{Sun_CVPR_2018}
Deqing Sun, Xiaodong Yang, Ming-Yu Liu, and Jan Kautz.
\newblock {PWC-Net}: {CNNs} for optical flow using pyramid, warping, and cost
  volume.
\newblock In {\em IEEE Conference on Computer Vision and Pattern Recognition},
  2018.

\bibitem{szegedy2016inception}
Christian Szegedy, Sergey Ioffe, Vincent Vanhoucke, and Alex Alemi.
\newblock Inception-v4, inception-resnet and the impact of residual connections
  on learning.
\newblock {\em arXiv preprint arXiv:1602.07261}, 2016.

\bibitem{szegedy2015going}
Christian Szegedy, Wei Liu, Yangqing Jia, Pierre Sermanet, Scott Reed, Dragomir
  Anguelov, Dumitru Erhan, Vincent Vanhoucke, and Andrew Rabinovich.
\newblock Going deeper with convolutions.
\newblock In {\em Proceedings of the IEEE conference on computer vision and
  pattern recognition}, pages 1--9, 2015.

\bibitem{tan2019mnasnet}
Mingxing Tan, Bo Chen, Ruoming Pang, Vijay Vasudevan, Mark Sandler, Andrew
  Howard, and Quoc~V Le.
\newblock Mnasnet: Platform-aware neural architecture search for mobile.
\newblock In {\em Proceedings of the IEEE Conference on Computer Vision and
  Pattern Recognition}, pages 2820--2828, 2019.

\bibitem{tan2019efficientnet}
Mingxing Tan and Quoc~V Le.
\newblock Efficientnet: Rethinking model scaling for convolutional neural
  networks.
\newblock {\em arXiv preprint arXiv:1905.11946}, 2019.

\bibitem{tao2017detail}
Xin Tao, Hongyun Gao, Renjie Liao, Jue Wang, and Jiaya Jia.
\newblock Detail-revealing deep video super-resolution.
\newblock In {\em Proceedings of the IEEE International Conference on Computer
  Vision}, pages 4472--4480, 2017.

\bibitem{tian2020tdan}
Yapeng Tian, Yulun Zhang, Yun Fu, and Chenliang Xu.
\newblock Tdan: Temporally-deformable alignment network for video
  super-resolution.
\newblock In {\em Proceedings of the IEEE/CVF Conference on Computer Vision and
  Pattern Recognition}, pages 3360--3369, 2020.

\bibitem{tong2017image}
Tong Tong, Gen Li, Xiejie Liu, and Qinquan Gao.
\newblock Image super-resolution using dense skip connections.
\newblock In {\em Proceedings of the IEEE International Conference on Computer
  Vision}, pages 4799--4807, 2017.

\bibitem{edvr2019wang}
Xintao Wang, Kelvin~CK Chan, Ke Yu, Chao Dong, and Chen Change~Loy.
\newblock Edvr: Video restoration with enhanced deformable convolutional
  networks.
\newblock In {\em Proceedings of the IEEE Conference on Computer Vision and
  Pattern Recognition Workshops}, pages 0--0, 2019.

\bibitem{wang2018esrgan}
Xintao Wang, Ke Yu, Shixiang Wu, Jinjin Gu, Yihao Liu, Chao Dong, Yu Qiao, and
  Chen Change~Loy.
\newblock Esrgan: Enhanced super-resolution generative adversarial networks.
\newblock In {\em Proceedings of the European Conference on Computer Vision
  (ECCV)}, pages 0--0, 2018.

\bibitem{wen2016learning}
Wei Wen, Chunpeng Wu, Yandan Wang, Yiran Chen, and Hai Li.
\newblock Learning structured sparsity in deep neural networks.
\newblock In {\em NIPS}, 2016.

\bibitem{wu2016quantized}
Jiaxiang Wu, Cong Leng, Yuhang Wang, Qinghao Hu, and Jian Cheng.
\newblock Quantized convolutional neural networks for mobile devices.
\newblock In {\em CVPR}, 2016.

\bibitem{xue2019video}
Tianfan Xue, Baian Chen, Jiajun Wu, Donglai Wei, and William~T Freeman.
\newblock Video enhancement with task-oriented flow.
\newblock {\em International Journal of Computer Vision}, 127(8):1106--1125,
  2019.

\bibitem{yim2017gift}
Junho Yim, Donggyu Joo, Jihoon Bae, and Junmo Kim.
\newblock A gift from knowledge distillation: Fast optimization, network
  minimization and transfer learning.
\newblock In {\em CVPR}, pages 4133--4141, 2017.

\bibitem{yu2018nisp}
Ruichi Yu, Ang Li, Chun-Fu Chen, Jui-Hsin Lai, Vlad~I Morariu, Xintong Han,
  Mingfei Gao, Ching-Yung Lin, and Larry~S Davis.
\newblock Nisp: Pruning networks using neuron importance score propagation.
\newblock In {\em Proceedings of the IEEE Conference on Computer Vision and
  Pattern Recognition}, pages 9194--9203, 2018.

\bibitem{yu2020joint}
Songhyun Yu, Bumjun Park, Junwoo Park, and Jechang Jeong.
\newblock Joint learning of blind video denoising and optical flow estimation.
\newblock In {\em Proceedings of the IEEE/CVF Conference on Computer Vision and
  Pattern Recognition Workshops}, pages 500--501, 2020.

\bibitem{zhang2017beyond}
Kai Zhang, Wangmeng Zuo, Yunjin Chen, Deyu Meng, and Lei Zhang.
\newblock Beyond a gaussian denoiser: Residual learning of deep cnn for image
  denoising.
\newblock {\em IEEE Transactions on Image Processing}, 26(7):3142--3155, 2017.

\bibitem{zoph2016neural}
Barret Zoph and Quoc~V Le.
\newblock Neural architecture search with reinforcement learning.
\newblock {\em arXiv preprint arXiv:1611.01578}, 2016.

\end{thebibliography}
}

\appendices

\section{Qualitative results on the Vimeo-90K dataset}
\label{ssec:qual_ablate}

\subsection{Deblocking}
\label{ssec:deblock_ablate}
\begin{figure*}[h!]
    \centering
    \begin{subfigure}[b]{2\columnwidth}
        \centering
        \includegraphics[width=0.49\columnwidth]{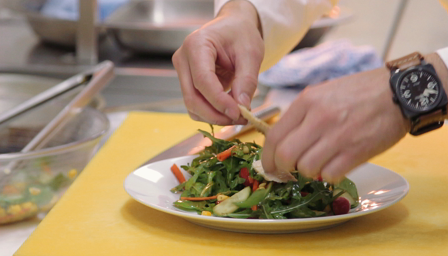}
        \caption{Original}
    \end{subfigure}
    \vfill
    \begin{subfigure}[b]{2\columnwidth}
        \centering
        \begin{tabular}{cc}
            \includegraphics[width=0.49\columnwidth]{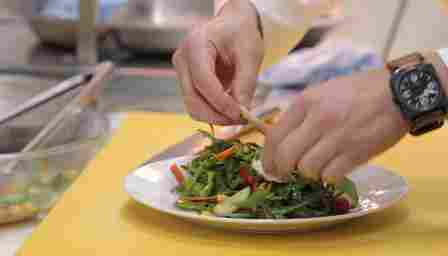} &  \includegraphics[width=0.49\columnwidth]{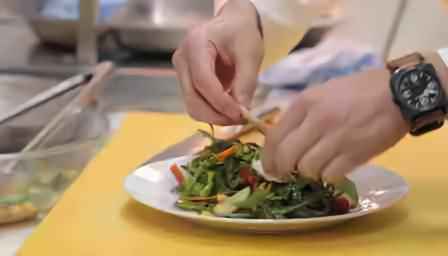}
        \end{tabular}
        \caption{\textbf{Left:} Compressed frame ($Q=15$). \textbf{Right:} Deblocked image (RGB PSNR: 31.31 dB)}
        \label{fig:ex1_q15}
    \end{subfigure}
    \vfill
    \begin{subfigure}[b]{2\columnwidth}
        \centering
        \begin{tabular}{cc}
            \includegraphics[width=0.49\columnwidth]{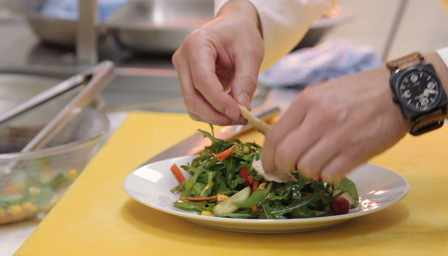} &  \includegraphics[width=0.49\columnwidth]{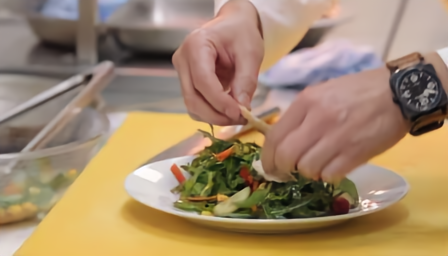}
        \end{tabular}
        \caption{\textbf{Left:} Compressed frame ($Q=45$). \textbf{Right:} Deblocked image (RGB PSNR: 34.79 dB)}
    \end{subfigure}
    \caption{Deblocking example at different values of $Q$. Note that lower value of $Q$ means higher compression.}
    \label{fig:deblock_example_a}
\end{figure*}

\begin{figure*}[h!]
    \centering
    \begin{subfigure}[b]{2\columnwidth}
        \centering
        \includegraphics[width=0.49\columnwidth]{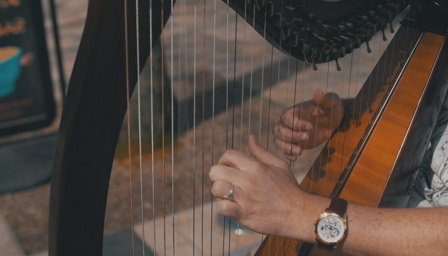}
        \caption{Original}
    \end{subfigure}
    \vfill
    \begin{subfigure}[b]{2\columnwidth}
        \centering
        \begin{tabular}{cc}
            \includegraphics[width=0.49\columnwidth]{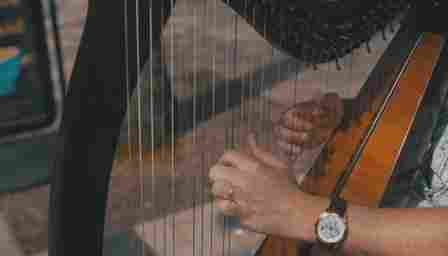} &  \includegraphics[width=0.49\columnwidth]{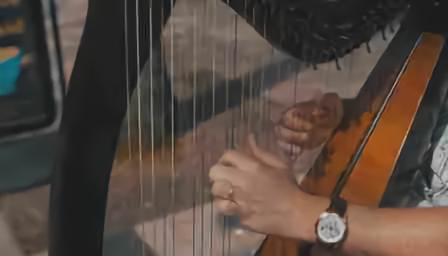}
        \end{tabular}
        \caption{\textbf{Left:} Compressed frame ($Q=15$). \textbf{Right:} Deblocked image (RGB PSNR: 32.11 dB)}
    \end{subfigure}
    \vfill
    \begin{subfigure}[b]{2\columnwidth}
        \centering
        \begin{tabular}{cc}
            \includegraphics[width=0.49\columnwidth]{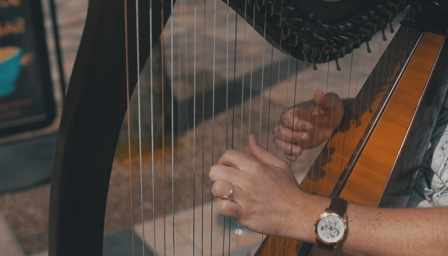} &  \includegraphics[width=0.49\columnwidth]{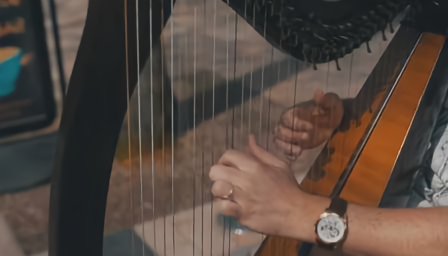}
        \end{tabular}
        \caption{\textbf{Left:} Compressed frame ($Q=45$). \textbf{Right:} Deblocked image (RGB PSNR: 36.21 dB)}
    \end{subfigure}
    \caption{Deblocking example at different values of $Q$. Note that lower value of $Q$ means higher compression.}
    \label{fig:deblock_example_b}
\end{figure*}

\begin{figure*}[h!]
    \centering
    \begin{subfigure}[b]{2\columnwidth}
        \centering
        \includegraphics[width=0.49\columnwidth]{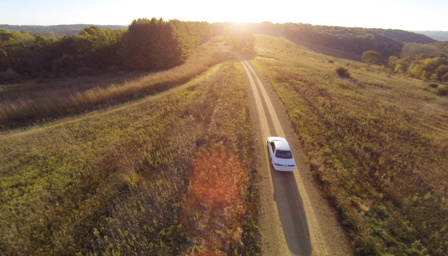}
        \caption{Original}
    \end{subfigure}
    \vfill
    \begin{subfigure}[b]{2\columnwidth}
        \centering
        \begin{tabular}{ccc}
            \includegraphics[width=0.49\columnwidth]{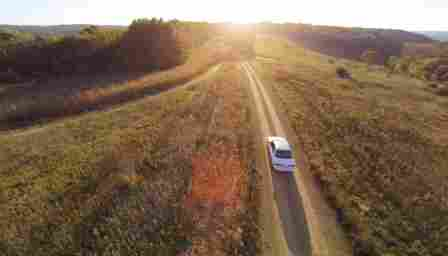} & \hfill &  \includegraphics[width=0.49\columnwidth]{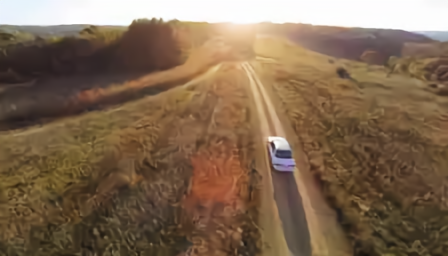}
        \end{tabular}
        \caption{\textbf{Left:} Compressed frame ($Q=15$). \textbf{Right:} Deblocked image (RGB PSNR: 30.23 dB)}
    \end{subfigure}
    \vfill
    \begin{subfigure}[b]{2\columnwidth}
        \centering
        \begin{tabular}{ccc}
            \includegraphics[width=0.49\columnwidth]{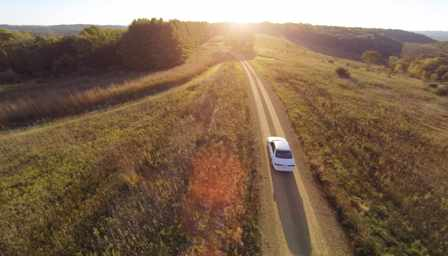} & \hfill &  \includegraphics[width=0.49\columnwidth]{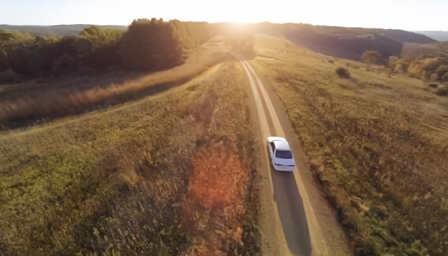}
        \end{tabular}
        \caption{\textbf{Left:} Compressed frame ($Q=45$). \textbf{Right:} Deblocked image (RGB PSNR: 33.02 dB)}
    \end{subfigure}
    \caption{Deblocking example at different values of $Q$. Note that lower value of $Q$ means higher compression.}
    \label{fig:deblock_example_c}
\end{figure*}

Figures \ref{fig:deblock_example_a}, \ref{fig:deblock_example_b}, and \ref{fig:deblock_example_c} demonstrate \arch's ability in deblocking videos at different compression factors in diverse environments ($Q$; lower value of $Q$ means higher compression). For example, in Figure \ref{fig:ex1_q15}, \arch~is able to remove the macro-block artifacts even under high compression ($Q=15$) around objects (e.g., hand, vegetables, and mixing bowl).

\subsection{Denoising}
\label{ssec:denoise_ablate}
\begin{figure*}[h!]
    \centering
    \begin{subfigure}[b]{2\columnwidth}
        \centering
        \begin{tabular}{ccc}
            \includegraphics[width=0.45\columnwidth]{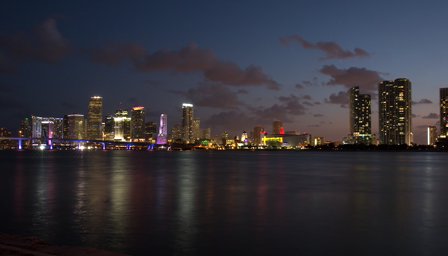} & \hfill &  \includegraphics[width=0.45\columnwidth]{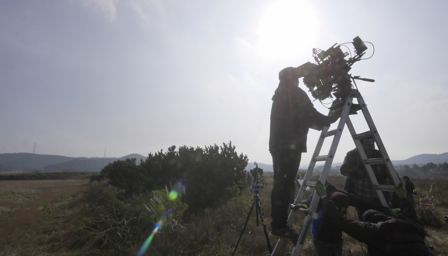}
        \end{tabular}
        \caption{Original frames}
    \end{subfigure}
    \vfill
    \begin{subfigure}[b]{2\columnwidth}
        \centering
        \begin{tabular}{ccc}
            \includegraphics[width=0.45\columnwidth]{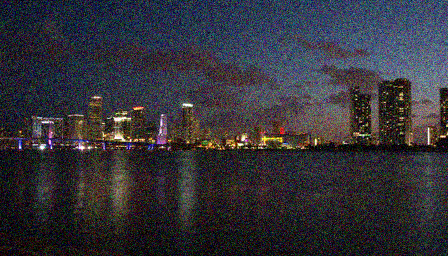} & \hfill &  \includegraphics[width=0.45\columnwidth]{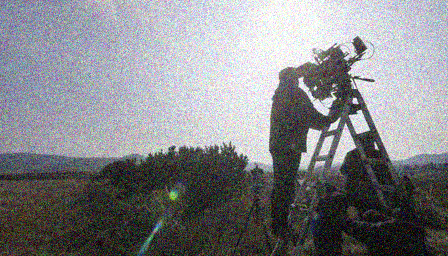} 
        \end{tabular}
        \caption{Noised frames with AWGN ($\sigma^2=0.01$).}
    \end{subfigure}
    \vfill
    \begin{subfigure}[b]{2\columnwidth}
        \centering
        \begin{tabular}{ccc}
            \includegraphics[width=0.45\columnwidth]{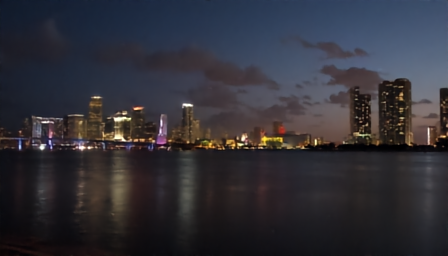} & \hfill & \includegraphics[width=0.45\columnwidth]{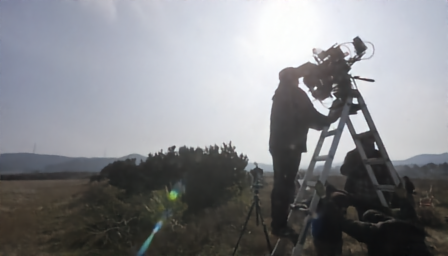}
        \end{tabular}
        \caption{Denoised frames with \arch. RGB PSNR of these denoised images is 33.94 dB (\textbf{left}) and 37.73 dB (\textbf{right}), respectively.}
        \label{fig:ex_awgn_high}
    \end{subfigure}
    \caption{AWGN denoising results on two different sequences.}
    \label{fig:awgn_example_a}
\end{figure*}

\begin{figure*}[t!]
    \centering
    \begin{subfigure}[b]{2\columnwidth}
        \centering
        \begin{tabular}{ccc}
            \includegraphics[width=0.45\columnwidth]{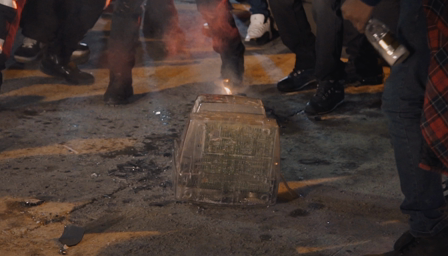} & \hfill & \includegraphics[width=0.45\columnwidth]{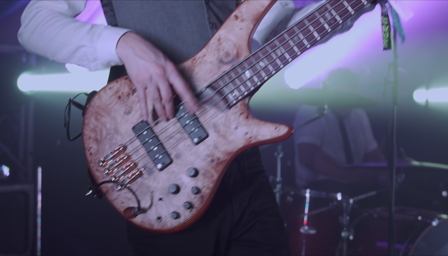}
        \end{tabular}
        \caption{Original}
    \end{subfigure}
    \vfill
    \begin{subfigure}[b]{2\columnwidth}
        \centering
        \begin{tabular}{ccc}
            \includegraphics[width=0.45\columnwidth]{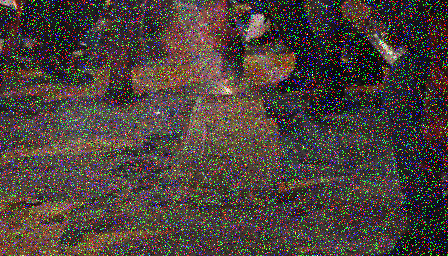} & \hfill &  \includegraphics[width=0.45\columnwidth]{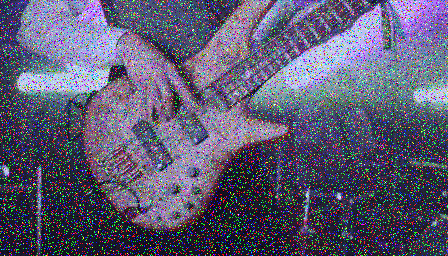}
        \end{tabular}
        \caption{Noised images with S\&P ($\rho=0.15$)}
    \end{subfigure}
    \vfill
    \begin{subfigure}[b]{2\columnwidth}
        \centering
        \begin{tabular}{ccc}
            \includegraphics[width=0.45\columnwidth]{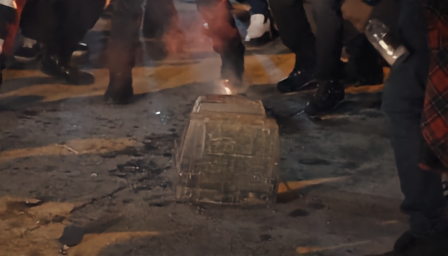} & \hfill & \includegraphics[width=0.45\columnwidth]{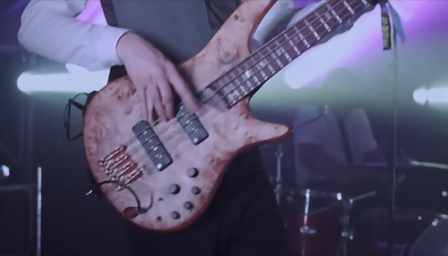}
        \end{tabular}
        \caption{Denoised images with \arch. RGB PSNR of denoised images is 37.28 dB and 35.50 dB, respectively.}
    \end{subfigure}
    \caption{Salt \& Pepper Denoising Example}
    \label{fig:snp_example_a}
\end{figure*}

\begin{figure*}[t!]
    \centering
    \resizebox{2\columnwidth}{!}{
    \begin{tabular}{ccc}
        \includegraphics[width=0.9\columnwidth]{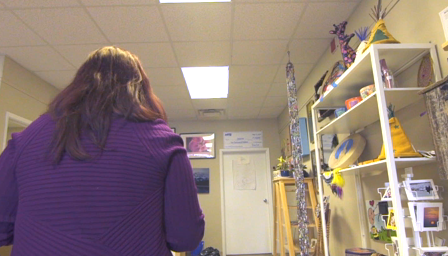} & \hfill & \includegraphics[width=0.9\columnwidth]{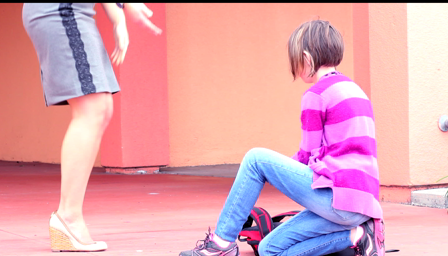} \\
        \multicolumn{3}{c}{(a) Original images} \\
        \includegraphics[width=0.9\columnwidth]{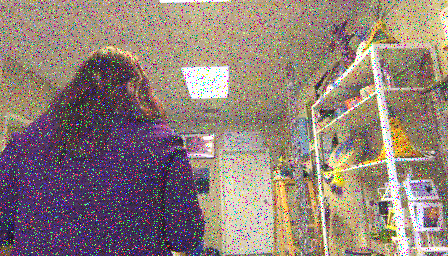} & \hfill & \includegraphics[width=0.9\columnwidth]{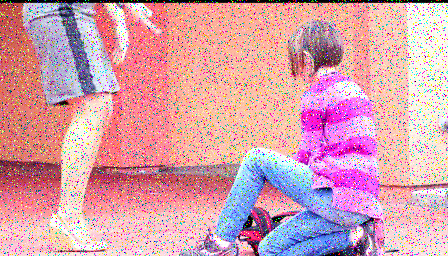} \\
        \multicolumn{3}{c}{(b) Noised images with AWGN ($\sigma^2=0.001$) and S\&P ($\rho=0.1$)} \\
        \includegraphics[width=0.9\columnwidth]{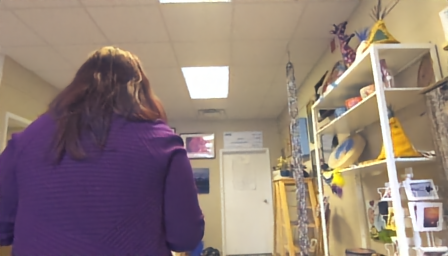} & \hfill & \includegraphics[width=0.9\columnwidth]{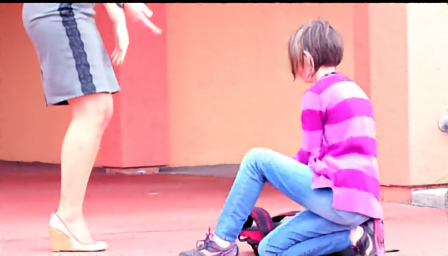} \\
        \multicolumn{3}{c}{(c) Denoised images with \arch. RGB PSNR of denoised images is 32.55 dB (\textbf{left}) and 32.09 dB (\textbf{right}), respectively.} \\
    \end{tabular}
    }
    \caption{Denoising example with mixed noise}
    \label{fig:mixed_noise}
\end{figure*}

Figures \ref{fig:awgn_example_a}, \ref{fig:snp_example_a}, and \ref{fig:mixed_noise} demonstrates \arch's ability in denoising different types of noise. For example, in Figure \ref{fig:ex_awgn_high}, \arch~is able to remove the noise and restore videos with high-quality.

\subsection{Video super-resolution ($4\times$)}
\label{ssec:vsr_ablate}
\begin{figure*}[h!]
    \centering
    \begin{subfigure}[b]{2\columnwidth}
        \centering
        \begin{tabular}{cc}
            \includegraphics[scale=0.5]{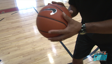} & \includegraphics[scale=0.5]{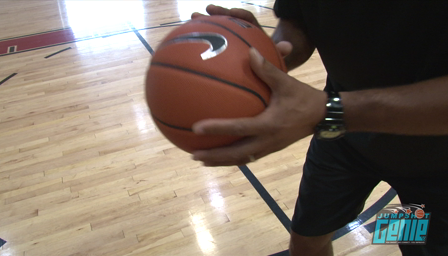} \\
            \includegraphics[scale=0.5]{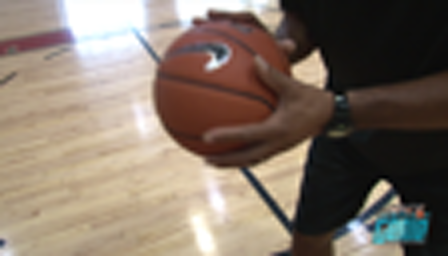} &
            \includegraphics[scale=0.5]{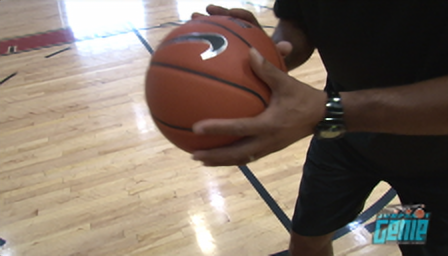} \\
            
        \end{tabular}
        \caption{\textbf{Top left:} Input low-resolution frame. \textbf{Top right:} Ground truth. \textbf{Bottom left:} Output of bicubic up-sampling (RGB PSNR: 28.59 dB) \textbf{Bottom right:} Output of \arch~(RGB PSNR=34.76 dB).}
    \end{subfigure}
    \vfill
    \begin{subfigure}[b]{2\columnwidth}
        \centering
        \begin{tabular}{cc}
            \includegraphics[scale=0.5]{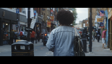} & \includegraphics[scale=0.5]{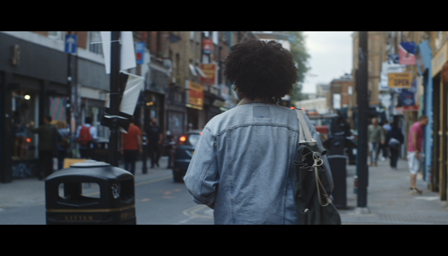} \\
            \includegraphics[scale=0.5]{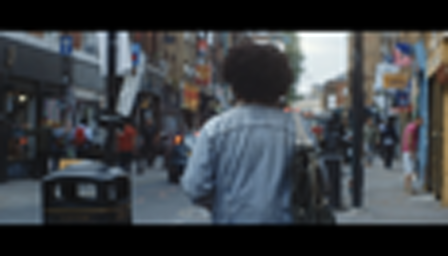} &
            \includegraphics[scale=0.5]{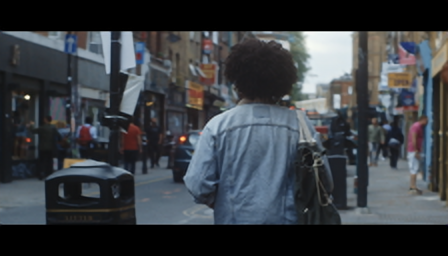} \\
            
        \end{tabular}
        \caption{\textbf{Top left:} Input low-resolution frame. \textbf{Top right:} Ground truth. \textbf{Bottom left:} Output of bicubic up-sampling (RGB PSNR: 27.56 dB) \textbf{Bottom right:} Output of \arch~(RGB PSNR=38.41 dB).}
    \end{subfigure}
    \caption{$4\times$ Video super-resolution examples.}
    \label{fig:video_sr_ex_a}
\end{figure*}

\begin{figure*}[h!]
    \centering
    \begin{subfigure}[b]{2\columnwidth}
        \centering
        \begin{tabular}{cc}
            \includegraphics[scale=0.5]{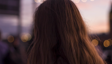} & \includegraphics[scale=0.5]{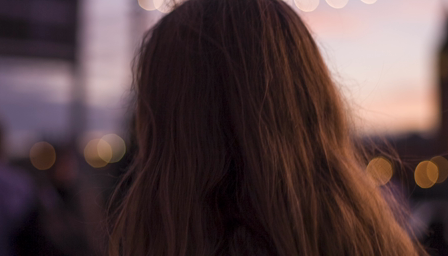} \\
            \includegraphics[scale=0.5]{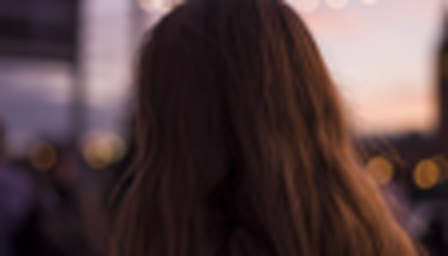} &
            \includegraphics[scale=0.5]{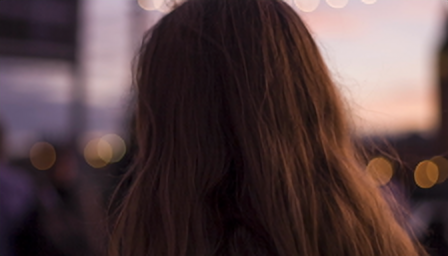} \\
            
        \end{tabular}
        \caption{\textbf{Top left:} Input low-resolution frame. \textbf{Top right:} Ground truth. \textbf{Bottom left:} Output of bicubic up-sampling (RGB PSNR: 36.84 dB) \textbf{Bottom right:} Output of \arch~(RGB PSNR=42.84 dB).}
        \label{fig:super_res_exa}
    \end{subfigure}
    \vfill
    \begin{subfigure}[b]{2\columnwidth}
        \centering
        \begin{tabular}{cc}
            \includegraphics[scale=0.5]{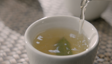} & \includegraphics[scale=0.5]{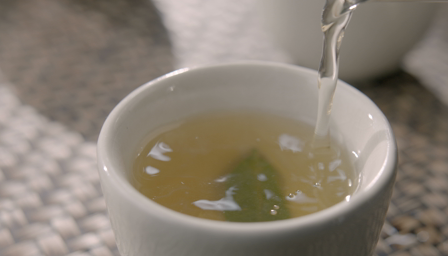} \\
            \includegraphics[scale=0.5]{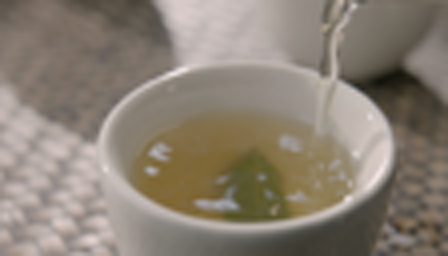} & 
            \includegraphics[scale=0.5]{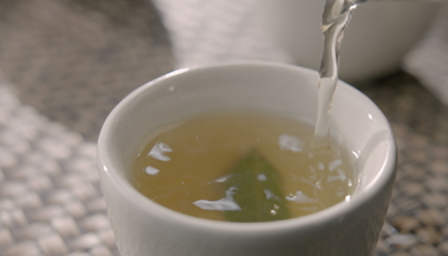}
        \end{tabular}
        \caption{\textbf{Top left:} Input low-resolution frame. \textbf{Top right:} Ground truth. \textbf{Bottom left:} Output of bicubic up-sampling (RGB PSNR: 36.21 dB) \textbf{Bottom right:} Output of \arch~(RGB PSNR=43.97 dB).}
    \end{subfigure}
    \caption{$4\times$ Video super-resolution examples.}
    \label{fig:video_sr_ex_b}
\end{figure*}

Figure \ref{fig:video_sr_ex_a} and \ref{fig:video_sr_ex_b} shows that \arch~is effective in restoring the details for $4\times$ video super-resolution. For example, in Figure \ref{fig:super_res_exa}, \arch~is able to restore fine details (e.g., hair strands) which are hard to restore with bicubic interpolation.

\section{Ablations}
\label{sec:ablations_append}

\noindent \textbf{Effect of different CUs:}  Table \ref{tab:cu_study} studies the effect of single- and multi-scale convolutional units (CUs) with and without SE unit. Multi-scale CU units with SE help improve the performance in case of AWGN denoising while no gain was observed in case of deblocking and super-resolution. We hypothesize that this is because compression happens at macro-block level, and both single and multi-scale blocks are able to effectively remove compression artifacts. Unlike macro-block compression, AWGN noise is identically distributed in the frames and kernels at different scales helps learn better representations and remove noisy artifacts (see gray color row in Table \ref{tab:block_ablate_denoising}).

\vspace{1mm}
\noindent \textbf{Effect of the depth of alignment, differential, and fusion modules:} Table \ref{tab:ablate_depth} studies \arch~with different values of $N_A$, $N_D$, and $N_F$. We are interested in efficient networks for edge devices, therefore, we studied only those combinations that satisfies this criteria: $N_A + N_D + N_F=9$. Similar to the effect of different CUs, we did not observe much gains when varying the depth of alignment, differential, and fusion modules for the task of deblocking and super-resolution. However, for denoising, we found that deeper alignment modules delivers the best trade-off between performance and MACs. Therefore, in our main experiments, we used $N_A=5$, $N_D=2$, and $N_F=2$ (see gray color row in Table \ref{tab:ablate_depth}).

\begin{table*}[h!]
\centering
\begin{subtable}[b]{2\columnwidth}
    \centering
    \resizebox{0.7\columnwidth}{!}{
    \begin{tabular}{lcrrrrcrr}
    \toprule[1.5pt]
            &         &         &            & \multicolumn{2}{c}{\bfseries RGB} && \multicolumn{2}{c}{\bfseries Y-Channel} \\
            \cmidrule[1pt]{5-6}\cmidrule[1pt]{8-9}
    {\bfseries CU Type} & {\bfseries SE Unit} & {\bfseries MACs}    & {\bfseries \# Params} & {\bfseries PSNR}        & {\bfseries SSIM}      && {\bfseries PSNR} & {\bfseries SSIM}         \\
    \midrule[1pt]
    Single  & \xmark       & 9.85 G  & 68.15 K    & 36.358      & 0.948     && 38.477         & 0.961        \\
    Single  & \cmark       & 9.85 G  & 72.95 K    & 36.323      & 0.948     && 38.403         & 0.961        \\
    \midrule
    Multi   & \xmark       & 10.79 G & 73.91 K    & 36.297      & 0.947     && 38.363         & 0.961        \\
    Multi   & \cmark       & 10.79 G & 78.71 K    & 36.334      & 0.948     && 38.478         & 0.962   \\
    \bottomrule[1.5pt]
    \end{tabular}
    }
    \caption{Deblocking ($Q = 40$)}
    \label{tab:block_ablate_deblock}
\end{subtable}
\hfill 
\begin{subtable}[b]{2\columnwidth}
    \centering
    \resizebox{0.7\columnwidth}{!}{
    \begin{tabular}{lcrrrrcrr}
    \toprule[1.5pt]
            &         &         &            & \multicolumn{2}{c}{\bfseries RGB} && \multicolumn{2}{c}{\bfseries Y-Channel} \\
            \cmidrule[1pt]{5-6}\cmidrule[1pt]{8-9}
    {\bfseries CU Type} & {\bfseries SE Unit} & {\bfseries MACs}    & {\bfseries \# Params} & {\bfseries PSNR}        & {\bfseries SSIM}      && {\bfseries PSNR} & {\bfseries SSIM}         \\
    \midrule[1pt]
    Single  & \xmark       & 9.85 G  & 68.15 K    & 31.207      & 0.868     && 32.650         & 0.886        \\
    Single  & \cmark       & 9.85 G  & 72.95 K    & 32.006      & 0.896     && 33.365         & 0.914        \\
    \midrule
    Multi   & \xmark       & 10.79 G & 73.91 K    & 29.026      & 0.875     && 30.247         & 0.895        \\
    \rowcolor{gray!30}
    Multi   & \cmark       & 10.79 G & 78.71 K    & 32.370      & 0.900     && 33.679         & 0.916       \\
    \bottomrule[1.5pt]
    \end{tabular}
    }
    \caption{AWGN Denoising ($\sigma^2 = 0.001$)}
    \label{tab:block_ablate_denoising}
\end{subtable}
\vfill
\begin{subtable}[b]{2\columnwidth}
\centering
    \resizebox{0.7\columnwidth}{!}{
    \begin{tabular}{lcrrrrcrr}
    \toprule[1.5pt]
            &         &         &            & \multicolumn{2}{c}{\bfseries RGB} && \multicolumn{2}{c}{\bfseries Y-Channel} \\
            \cmidrule[1pt]{5-6}\cmidrule[1pt]{8-9}
    {\bfseries CU Type} & {\bfseries SE Unit} & {\bfseries MACs}    & {\bfseries \# Params} & {\bfseries PSNR}        & {\bfseries SSIM}      && {\bfseries PSNR} & {\bfseries SSIM}         \\
    \midrule[1pt]
    Single  & \xmark       & 9.90 G  & 68.33 K    & 37.406      & 0.962     && 38.042         & 0.966        \\
    Single  & \cmark       & 9.90 G  & 73.14 K    & 37.318      & 0.962     && 37.955         & 0.965        \\
    Multi   & \xmark       & 10.84 G & 74.10 K    & 37.181      & 0.962     && 37.868         & 0.966        \\
    Multi   & \cmark       & 10.84 G & 78.91 K    & 37.378      & 0.962     && 38.002         & 0.966     \\
    \bottomrule[1.5pt]
    \end{tabular}
    }
    \caption{Super-resolution ($2\times$)}
    \label{tab:block_ablate_superres}
\end{subtable}
\caption{\textbf{Effect of different CU units.} Multi-scale blocks are effective in restoring fine-grained details (e.g., noise) while both single- and multi-scale blocks are effective in restoring block-level artifacts (e.g., compression). Here, we used $N_A=N_D=N_F=3$.}
\label{tab:cu_study}
\end{table*}

\begin{table*}[h!]
\centering
\begin{subtable}[b]{2\columnwidth}
    \centering
    \resizebox{0.7\columnwidth}{!}{
    \begin{tabular}{cccrrrrcrr}
    \toprule[1.5pt]
    \multicolumn{3}{c}{\bfseries Module depth} & & & \multicolumn{2}{c}{\bfseries RGB} && \multicolumn{2}{c}{\bfseries Y-Channel} \\
    \cmidrule[1pt]{1-3}\cmidrule[1pt]{6-7}\cmidrule[1pt]{9-10}
    $N_A$ & $N_D$ & $N_F$ & {\bfseries MACs}    & {\bfseries \# Params} & {\bfseries PSNR}        & {\bfseries SSIM}      && {\bfseries PSNR} & {\bfseries SSIM}  \\
    \midrule[1.25pt]
    1    & 1    & 7    & 11.44 G & 78.71 K    & 36.320 & 0.948 && 38.411 & 0.961 \\
    1    & 7    & 1    & 11.44 G & 78.71 K    & 36.356 & 0.948 && 38.450 & 0.962 \\
    7    & 1    & 1    & 9.47 G  & 78.71 K    & 36.334 & 0.948 && 38.472 & 0.961 \\
    \midrule
    2    & 2    & 5    & 11.11 G & 78.71 K    & 36.200 & 0.946 && 38.297 & 0.960 \\
    2    & 5    & 2    & 11.11 G & 78.71 K    & 36.327 & 0.948 && 38.412 & 0.962 \\
    5    & 2    & 2    & 10.13 G & 78.71 K    & 36.307 & 0.947 && 38.403 & 0.961 \\
    \midrule
    3    & 2    & 4    & 10.77 G & 78.71 K    & 36.359 & 0.948 && 38.451 & 0.962 \\
    3    & 4    & 2    & 10.77 G & 78.71 K    & 36.307 & 0.947 && 38.390 & 0.961 \\
    4    & 3    & 2    & 10.46 G & 78.71 K    & 36.287 & 0.948 && 38.405 & 0.961 \\
    \midrule
    3    & 3    & 3    & 10.79 G & 78.71 K    & 36.334 & 0.948 && 38.478 & 0.962 \\
    \bottomrule[1.5pt]
    \end{tabular}
    }
    \caption{Deblocking ($Q=40$)}
    \label{tab:ablate_depth_debl}
\end{subtable}
\vfill
\begin{subtable}[b]{2\columnwidth}
    \centering
    \resizebox{0.7\columnwidth}{!}{
    \begin{tabular}{cccrrrrcrr}
    \toprule[1.5pt]
    \multicolumn{3}{c}{\bfseries Module depth} & & & \multicolumn{2}{c}{\bfseries RGB} && \multicolumn{2}{c}{\bfseries Y-Channel} \\
    \cmidrule[1pt]{1-3}\cmidrule[1pt]{6-7}\cmidrule[1pt]{9-10}
    $N_A$ & $N_D$ & $N_F$ & {\bfseries MACs}    & {\bfseries \# Params} & {\bfseries PSNR}        & {\bfseries SSIM}      && {\bfseries PSNR} & {\bfseries SSIM}  \\
    \midrule[1.25pt]
    1 & 1 & 7 & 11.44 G & 78.71 K & 31.605 & 0.887 && 32.913 & 0.905 \\
1 & 7 & 1 & 11.44 G & 78.71 K & 31.753 & 0.884 && 32.951 & 0.901 \\
7 & 1 & 1 & 9.47 G  & 78.71 K & 30.859 & 0.871 && 32.139 & 0.890 \\
\midrule
2 & 2 & 5 & 11.11 G & 78.71 K & 32.139 & 0.901 && 33.477 & 0.919 \\
2 & 5 & 2 & 11.11 G & 78.71 K & 32.057 & 0.891 && 33.445 & 0.908 \\
\rowcolor{gray!30}
5 & 2 & 2 & 10.13 G & 78.71 K & 32.403 & 0.903 && 33.884 & 0.921 \\
\midrule
3 & 2 & 4 & 10.77 G & 78.71 K & 31.690 & 0.890 && 33.047 & 0.908 \\
3 & 4 & 2 & 10.77 G & 78.71 K & 30.785 & 0.874 && 32.193 & 0.896 \\
4 & 3 & 2 & 10.46 G & 78.71 K & 31.416 & 0.877 && 32.690 & 0.895 \\
\midrule
3 & 3 & 3 & 10.79 G & 78.71 K & 32.370 & 0.900 && 33.679 & 0.916 \\
    \bottomrule[1.5pt]
    \end{tabular}
    }
    \caption{AWGN Denoising ($\sigma^2 = 0.001$)}
    \label{tab:ablate_depth_denoi}
\end{subtable}
\vfill
\begin{subtable}[b]{2\columnwidth}
    \centering
    \resizebox{0.7\columnwidth}{!}{
    \begin{tabular}{cccrrrrcrr}
    \toprule[1.5pt]
    \multicolumn{3}{c}{\bfseries Module depth} & & & \multicolumn{2}{c}{\bfseries RGB} && \multicolumn{2}{c}{\bfseries Y-Channel} \\
    \cmidrule[1pt]{1-3}\cmidrule[1pt]{6-7}\cmidrule[1pt]{9-10}
    $N_A$ & $N_D$ & $N_F$ & {\bfseries MACs}    & {\bfseries \# Params} & {\bfseries PSNR}        & {\bfseries SSIM}      && {\bfseries PSNR} & {\bfseries SSIM}  \\
    \midrule[1.25pt]
    1 & 1 & 7 & 11.50 G & 78.91 K & 37.071 & 0.961 && 37.742 & 0.965 \\
1 & 7 & 1 & 11.50 G & 78.91 K & 37.136 & 0.961 && 37.774 & 0.965 \\
7 & 1 & 1 & 9.52 G  & 78.91 K & 37.176 & 0.961 && 37.868 & 0.965 \\
 \midrule
2 & 2 & 5 & 11.17 G & 78.91 K & 37.072 & 0.961 && 37.740 & 0.965 \\
2 & 5 & 2 & 11.17 G & 78.91 K & 37.102 & 0.961 && 37.776 & 0.965 \\
5 & 2 & 2 & 10.18 G & 78.91 K & 37.196 & 0.961 && 37.855 & 0.965 \\
\midrule
3 & 2 & 4 & 10.84 G & 78.91 K & 37.227 & 0.962 && 37.902 & 0.965 \\
3 & 4 & 2 &  10.84 G  & 78.91 K & 37.071 & 0.961 && 37.740 & 0.965 \\
4 & 3 & 2 & 10.51 G & 78.91 K & 37.173 & 0.961 && 37.877 & 0.965 \\
\midrule
3 & 3 & 3 & 10.84 G & 78.91 K & 37.378 & 0.962 && 38.002 & 0.966 \\
    \bottomrule[1.5pt]
    \end{tabular}
    }
    \caption{Super-resolution ($2\times$)}
    \label{tab:ablate_depth_sr}
\end{subtable}
\caption{\textbf{Effect of the depth of alignment, differential, and fusion modules in the \arch.} Overall, \arch~with deeper alignment modules provides the best trade-off between performance and number of multiplication-addition operations (MACs). In all these models, the depth of the network is fixed, i.e., $N_A + N_D + N_F = 9$. }
\label{tab:ablate_depth}
\end{table*}

\end{document}